\documentclass[lettersize,journal]{IEEEtran}
\usepackage{amsmath,amsfonts}
\usepackage{algorithmic}
\usepackage{algorithm}
\usepackage{array}
\usepackage[caption=false,font=normalsize,labelfont=sf,textfont=sf]{subfig}
\usepackage{textcomp}
\usepackage{stfloats}
\usepackage{url}
\usepackage{verbatim}
\usepackage{graphicx}
\usepackage[table,xcdraw]{xcolor}
\usepackage{multirow} 
\usepackage{booktabs} 
\usepackage{cite}
\usepackage[colorlinks,
            linkcolor=blue,
            anchorcolor=blue,
            citecolor=blue]{hyperref}
\usepackage{amssymb}
\usepackage{algorithm}
\usepackage{algorithmic}
\usepackage{orcidlink}  

\begin{document}
\title{MAFS: Masked Autoencoder for Infrared-Visible Image Fusion and Semantic Segmentation}

\author{Liying Wang\raisebox{1ex}{\orcidlink{0009-0006-2399-5823}}, Xiaoli Zhang\raisebox{1ex}{\orcidlink{0000-0001-8412-4956}},~\IEEEmembership{Member,~IEEE}, Chuanmin Jia\raisebox{1ex}{\orcidlink{0000-0002-7418-6245}},~\IEEEmembership{Member,~IEEE}, and Siwei Ma\raisebox{1ex}{\orcidlink{0000-0002-2731-5403}},~\IEEEmembership{Fellow,~IEEE}
\thanks{Corresponding authors: Xiaoli Zhang and Chuanmin Jia.}
\thanks{Liying Wang and Xiaoli Zhang are with the Key Laboratory of Symbolic Computation and Knowledge Engineering of Ministry of Education, Jilin University, Changchun 130012, China.}
\thanks{Chuanmin Jia is with the Wangxuan Institute of Computer Technology, Peking
University, Beijing 100871, China.}
\thanks{Siwei Ma is with the National
Engineering Research Center of Visual Technology, School of Computer Science, Peking University, Beijing 100871, China.}}

\markboth{Journal of \LaTeX\ Class Files,~Vol.~14, No.~8, August~2021}%
{Shell \MakeLowercase{\textit{et al.}}: A Sample Article Using IEEEtran.cls for IEEE Journals}


\maketitle

\begin{abstract}

Infrared-visible image fusion methods aim at generating fused images with good visual quality and also facilitate the performance of high-level tasks. Indeed, existing semantic-driven methods have considered semantic information injection for downstream applications. However, none of them investigates the potential for reciprocal promotion between pixel-wise image fusion and cross-modal feature fusion perception tasks from a macroscopic task-level perspective. To address this limitation, we propose a unified network for image fusion and semantic segmentation. MAFS is a parallel structure, containing a fusion sub-network and a segmentation sub-network. On the one hand, We devise a heterogeneous feature fusion strategy to enhance semantic-aware capabilities for image fusion. On the other hand, by cascading the fusion sub-network and a segmentation backbone, segmentation-related knowledge is transferred to promote feature-level fusion-based segmentation.
Within the framework, we design a novel multi-stage Transformer decoder to aggregate fine-grained multi-scale fused features efficiently. Additionally, a dynamic factor based on the max-min fairness allocation principle is introduced to generate adaptive weights of two tasks and guarantee smooth training in a multi-task manner. 
Extensive experiments demonstrate that our approach achieves competitive results compared with state-of-the-art methods. The code is available at \url{https://github.com/Abraham-Einstein/MAFS/}.
\end{abstract}


\section{Introduction}
\IEEEPARstart{w}{ith} the rapid development of optical instruments and hardware equipment, scene parsing\cite{PIG}, \cite{zhao2017pyramid} plays an important role in the autonomous driving and robot sensing field. The images captured by visual imaging sensors and infrared cameras possess their own unique properties. Visible imaging sensors excel at capturing detailed, high-resolution, color images that align with human perception\cite{ma2023overview}, making them ideal for applications needing precise detail and color accuracy. However, they depend heavily on ambient light, perform poorly in low-light conditions, and cannot penetrate fog, smoke, or dust. In contrast, infrared cameras can operate effectively in total darkness and detect heat signatures, making them useful for low-visibility environments, thermal detection, and challenging outdoor conditions. But they have limitations, including lower resolution and monochromatic imaging, which may reduce clarity and detail compared to visible imaging. Infrared-visible image fusion\cite{liu2024coconet}, \cite{li2023lrrnet}
endows with complementary characteristics of RGB and thermal images making it an essential technique for various applications, such as object detection\cite{M2FNet}, pedestrian re-identification\cite{peng2023adaptive}, tracking\cite{li2018cross}, and semantic segmentation\cite{tang2023rethinking}.

In the past decades, numerous traditional fusion techniques \cite{liu2016image}, \cite{zhao2020bayesian} have been proposed. These methods unavoidably have several disadvantages: 1) requiring complex hand-crafted fusion rules and 2) seldom considering the characteristics of different modalities. In deep-learning-based schemes, the general image fusion methods\cite{liang2022fusion}, \cite{xu2020u2fusion}, transformer-based methods\cite{ma2022swinfusion}, \cite{AFT}, and diffusion-based methods\cite{Dif-Fusion}, \cite{zhao2023ddfm} emphasize preserving the thermal targets and texture structures and also achieve satisfactory outcomes. Nevertheless, in terms of infrared-visible image fusion tasks, these solutions exclusively focus on the issue of image fusion and rarely exploit the correlation with downstream tasks.

\begin{figure}[h]
    \centering
    \includegraphics[width=\linewidth]{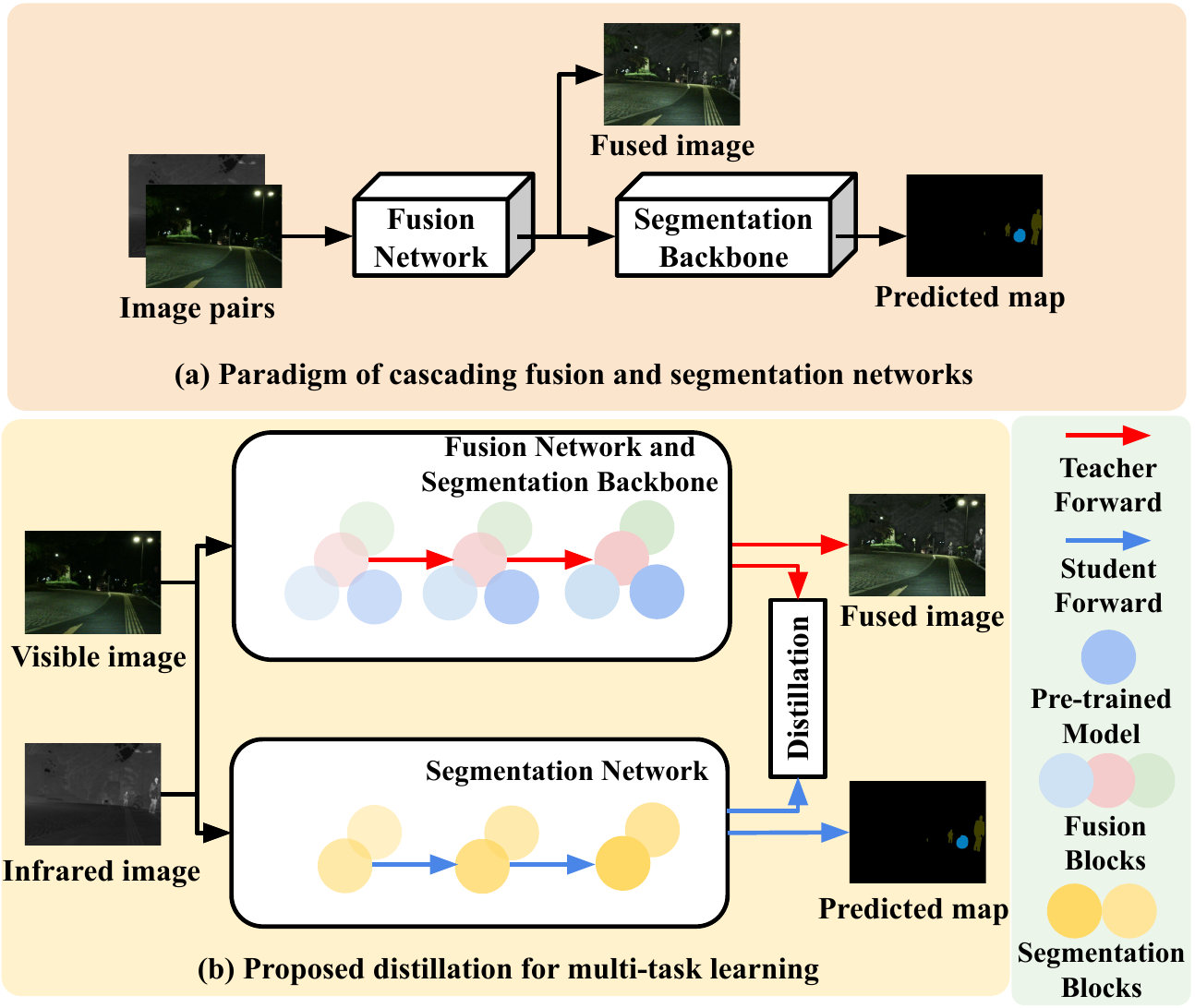}
    \caption{The joint training scheme. Previous frameworks cascade the fusion network and segmentation task, which brings performance limitations. We establish a teacher model comprising a fusion network and a segmentation backbone. The student model is a tailored segmentation network and its structure is different from the proposed teacher model.}
    \label{pipeline}
\end{figure}

Recently, researchers have begun to explore the multi-task learning domain. In 2021, Li \textit{et al.}\cite{Different} formulated a meta-learning manner via training infrared-visible image fusion and super-resolution simultaneously. The framework can accommodate source images of varying resolutions and produce fused images at any desired resolution. In 2022, Tang \textit{et al.}\cite{tang2022image} proposed a semantic-aware image fusion network. Encouragingly, this work first bridges the gap between image fusion and high-level tasks. Subsequently, DetFusion\cite{sun2022detfusion}, as a detection-driven network, utilized object-related information learned from detection networks to guide the process of image fusion. Typically, TarDal\cite{liu2022target} utilized bilevel optimization for the joint learning of fusion and detection in a target-aware adversarial network. In 2023, Liu \textit{et al.}\cite{liu2023multi} designed a multi-interactive learning pipeline in which the semantic and pixel-based features can achieve mutual interaction using a sequential connection. PAIF\cite{liu2023paif} is a perception-aware framework proposed to improve segmentation robustness when suffering from adversarial attacks. BDLFusion\cite{liu2023bilevel} employed the first-order approximation to approximate the Hessian matrix and utilized Random Loss Weighting (RLW) to adaptively aggregate task gradients. These methods all have considered the significance of task-oriented image fusion. By cascading the fusion network and the detection/segmentation network, the high-level tasks receive the fused images as the input for optimizing the fusion model and high-level task model shown in Fig. \hyperref[pipeline]{1(a)}, while this optimizing strategy may pose the potential risk of degraded performance in both tasks. 
Additionally, the difficulty of training different tasks during the optimization procedure remains to be further explored. As for the coupled learning frameworks, in 2023, PSFusion\cite{tang2023rethinking} guided task-specific feature learning via progressive semantic injection and scene fidelity constraints to improve the visual performance and task adaptation of image fusion. It demonstrated the potential of pixel-wise image fusion toward downstream tasks compared to cross-modal feature fusion methods. We define that the \textbf{microscopic network architecture perspective} achieves task-specific feature learning through the design of interactive modules and the \textbf{macroscopic task-level perspective} pursues mutual promotion among tasks. 
To be specific, advanced segmentation task-driven fusion methods implement semantic-aware fusion from a \textbf{microscopic network architecture perspective}. Different from the existing segmentation task-driven fusion methods, we not only leverage the advantages of the cascading paradigm to mitigate interference in the training process of image fusion but also consider the knowledge transferred from the image fusion to the specific high-level task to bridge their semantic gap from a \textbf{macroscopic task-level perspective}. 

To remedy the above issues, we propose a method for fusion and segmentation in a unified framework through knowledge distillation, as illustrated in Fig. \hyperref[pipeline]{1(b)}. When the {\itshape ``teacher''} is training the fusion network, it meanwhile guides the {\itshape ``student''} to achieve more stable segmentation network training based on the consistent constraints between their logits maps. Specifically, we devise a \textbf{M}asked \textbf{A}utoencoder for infrared-visible image \textbf{F}usion and semantic \textbf{S}egmentation (MAFS).
The multi-modal masked autoencoder is deployed into the two-stage training strategy as a pretext task for collaborative training. Furthermore, in the unified framework, the shallow feature extraction module (SFM) and progressive heterogeneous feature (PHF) fusion module aim to obtain consistent features about source images and fuse heterogeneous features. The fusion decoder comprises multiple dense invertible neural units to reduce the detail information loss. A dynamic extraction module (DEM) bridges the gap between the backbone network and the subsequent cross-attention fusion module (CAM) for better segmentation results. The segmentation decoder is based on the multi-stage transformer (MST), modeling global context information. To summarize, the main contributions of this work are organized into four aspects:
\begin{itemize}
    \item We introduce the knowledge distillation paradigm into a unified framework for image fusion and segmentation, aiming at exploring the relationship between pixel-wise image fusion and feature-level fusion in the context of multi-modal semantic segmentation.
    \item We employ the masked autoencoder for the reconstruction stage as a self-supervised way to capture abundant feature representation and initialize the second-stage training.
    \item We devise an adaptive and heterogeneous feature fusion strategy to achieve semantic-aware image fusion. Multilabel supervision and context information aggregation improve the precision of semantic segmentation.
    \item Our approach has demonstrated competitive results in infrared-visible image fusion and semantic segmentation compared with state-of-the-art methods.
\end{itemize}

Subsequently, in Section \ref{sec2}, we review some representative works of image fusion, along with some related works on semantic segmentation. Section \ref{sec3} describes the proposed MAFS model in detail. The experimental results and ablation studies can be found in Section \ref{sec4}. Section \ref{sec5} draws a concise summary.

\section{Related Works}
\label{sec2}
In this part, we briefly overview the development process of deep learning-based fusion methods and the evolution of semantic segmentation. 
\subsection{Infrared-Visible Image Fusion}
With the development of neural networks, deep learning has been applied to the infrared-visible image fusion field and occupies the dominant position. To sum up, these methods are mainly divided into three types: CNN-based methods, GAN-based methods, and AE-based methods. 

For CNN-based methods, rethinking the image fusion, PMGI\cite{zhang2020rethinking} first converted the fusion problem into proportional maintenance of intensity and gradient, which realized a variety of image fusion tasks. Subsequently, Zhang \textit{et al.}\cite{zhang2021sdnet} extracted and reconstructed the intensity and gradient information based on the loss function composed of intensity and gradient term. This squeeze-and-composition network enables handling a series of real-time fusion tasks. Nowadays, CNN-based methods also combine with transformers to enhance the capabilities of modeling long-range dependencies. To give an example, CHITNet\cite{CHITNet} achieved a transformer-based cross-modality mutual representation and transferred complementary features into a harmonious one.

For GAN-based methods, Ma \textit{et al.}\cite{ma2019fusiongan} designed the generator and the discriminator based on the convolutional neural network. FusionGAN formulated an adversarial game in order to preserve visible texture information and also keep infrared intensity information. 
Liu \textit{et al.}\cite{liu2021learning} proposed a deep feature ensemble network with an edge-guided attention mechanism as the generator and the subsequent discriminator guided the generator to reconstruct more natural images. These GAN-based methods suffer from mode collapse and instability in the training process. Diffusion-based models have shown their outstanding performance as generative methods. 
Dif-Fusion\cite{Dif-Fusion} learned the latent structure of multi-channel images while training the forward and reverse processes.
Yi \textit{et al.}\cite{yi2024diff} incorporated fusion knowledge prior into the forward process and designed the reverse diffusion process to generate fused images.


For AE-based methods, DenseFuse\cite{li2018densefuse} first proposed a novel network with an encoder and decoder. As a precursor to autoencoder-based deep learning methods, the encoding network includes convolutional layers and dense blocks, while the fusion strategy and the decoding network with four CNN layers reconstruct the fused image. 
Remarkably, Zhao \textit{et al.}\cite{zhao2023cddfuse} proposed a correlation-driven feature decomposition network to acquire modality-shared (common) and modality-specific (unique) features and further model cross-modality features. Due to the lack of ground truth, EMMA\cite{zhao2024equivariant}, as an end-to-end self-supervised learning paradigm, utilized the equivariant priors to follow the physical sensor imaging process. 


In consideration of the data compatibility\cite{liu2025} issue on misaligned infrared-visible image fusion, Li \textit{et al.}\cite{li2024deep} proposed a dynamic network, which embeds the alignment of source images into the feature extraction procedure. By introducing a learnable modality dictionary, MuIFS-CAP\cite{MulFS-CAP} proposed the cross-modality alignment mechanism via a correlation matrix, enabling the alignment and fusion in a single stage. Besides, the real-world application-oriented\cite{liu2025} approaches are also under the spotlight. Tang \textit{et al.}\cite{tang2022superfusion} proposed a versatile network that models image registration and fusion with semantic requirements. 
IRFS\cite{wang2023interactively} first explored the collaborative paradigm between image fusion and the salient object detection task. 
Subsequently,
Yang \textit{et al.}\cite{yang2025instruction} introduced dynamic prompt injection and extended image fusion into different task scenarios, i.e., object detection, semantic segmentation, and salient object detection. 
CAF\cite{CAF}, as a compact, automatic, and flexible framework, has novelly proposed the automatic searching for optimal fusion objectives. TIMFusion\cite{TIMFusion} developed the pretext meta initialization for the fast scenario adaptation, in which the implicit architecture search aims at efficient image fusion combined with task-related guidance. 
Recently, SAGE\cite{wu2025every} distilled semantic priors from the Segment Anything Model\cite{kirillov2023segment} to build up downstream task adaptation.

Inspired by both cascade-based and collaborative training approaches, we aim to further explore the potential of jointly training image fusion and semantic segmentation within the multi-task learning paradigm.


\subsection{Unimodal Semantic Segmentation}
DeepLabV3+\cite{chen2018encoder} injected the depthwise separable convolution into both Atrous Spatial Pyramid Pooling (ASPP) and the decoder module to achieve more fine-grained feature representation. 
High-Resolution Network (HRNet)\cite{wang2020deep} parallelly connected high-to-low resolution streams to obtain precise spatial representation and rich semantic information. 
It's worth pointing out that these methods put much effort into increasing the receptive field but ignored the significance of context modeling.
In transformer-based networks, to enhance global context information modeling, SETR\cite{zheng2021rethinking} treated semantic segmentation from a sequence-to-sequence perspective, which utilized a pure transformer to replace the encoder with convolution and resolution reduction. Based on the hierarchical Transformer paradigm, SegFormer\cite{xie2021segformer} removed positional encoding and also redesigned a simple MLP-based decoder, which had the benefit of smaller parameters and superior behavior. 

Given the methods’ performance that requires improvement, as mentioned above in complex environments, some researchers have shifted their focus to multimodal semantic segmentation to tackle challenging scenarios more effectively.
\subsection{Infrared-Visible Semantic Segmentation}
Due to the low-light environment at night and adverse weather conditions, MFNet\cite{ha2017mfnet} was a representative work for multi-spectral scene semantic segmentation towards autonomous driving, which adopted a lightweight encoder-decoder architecture. Unlike fusing outputs of encoders in the decoder step, RTFNet\cite{sun2019rtfnet} employed ResNet for feature extraction and fused infrared and visible information through element-wise summation. Shivakumar \textit{et al.}\cite{shivakumar2020pst900} proposed the PST900 dataset by leveraging long-wave infrared imagery and designed a dual-stream CNN-based architecture. These are earlier studies that exploit the advantages of modal complementarity. Considering modality differences owing to distinct imaging mechanisms, Zhang \textit{et al.}\cite{zhang2021abmdrnet} posed a \textit{bridging-then-fusing strategy} as the replacement for basic fusion strategies, e.g. element-wise summation and concatenation. GMNet\cite{zhou2021gmnet}, LASNet\cite{li2022rgb},  MMSMCNet\cite{zhou2023mmsmcnet}, EGFNet\cite{dong2023egfnet} made further investigations into multilabel supervision to capture the binary, boundary, and semantic characteristics during optimizing procedure. Recently, RoadFormer+\cite{li2024roadformer}, \cite{huang2024roadformer+} utilized learnable attention fusion modules from processing heterogeneous features perspective. These Transformer-based methods presented great potential for capturing both the correlation and differences between visible images and additional data types.
\begin{figure*}[h]
    \centering
    \includegraphics[width=\textwidth]{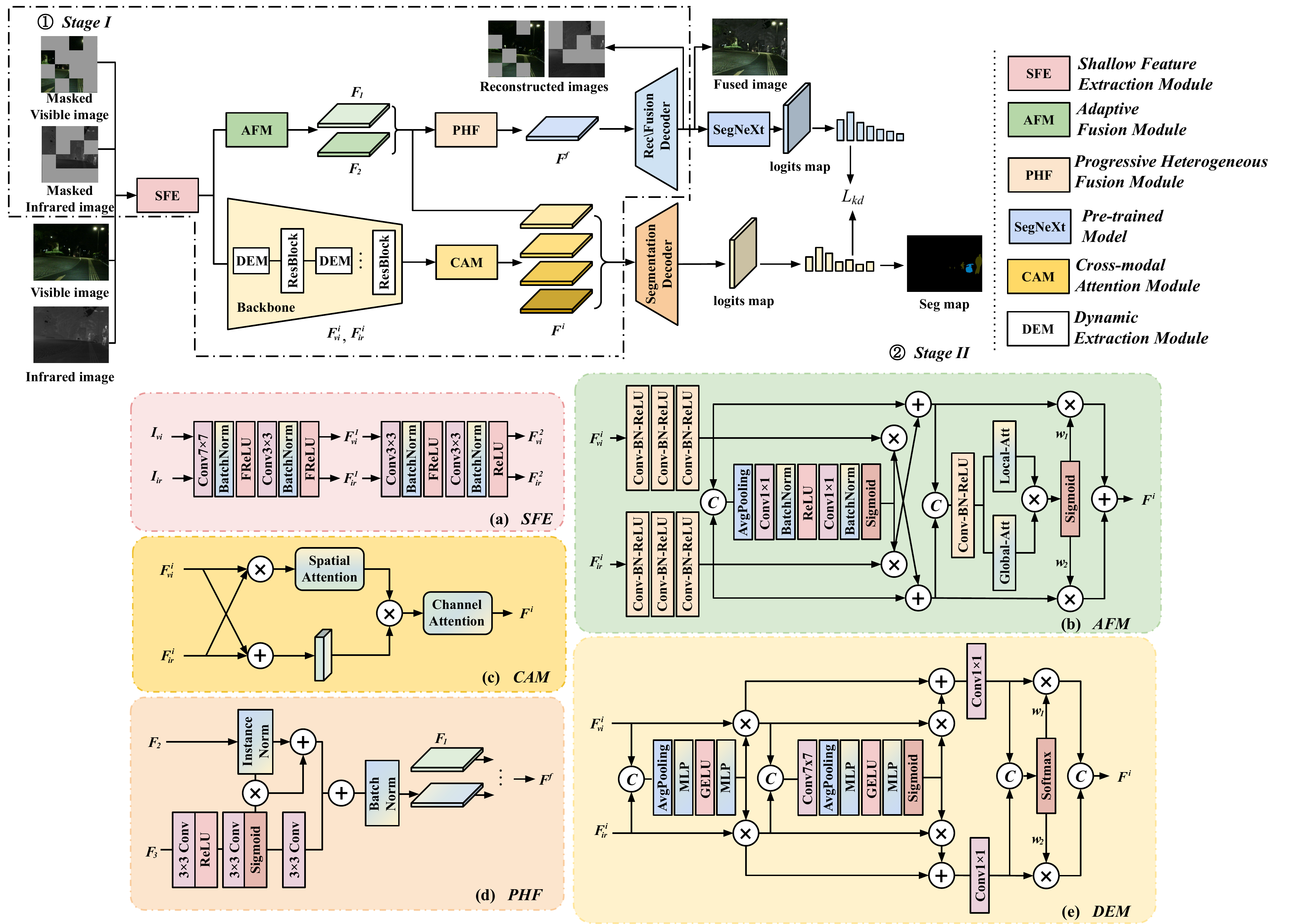}
    \caption{The detailed architectures are mainly three components: encoder, fusion strategies, and decoders. First, the shallow features and multi-scale features are extracted for the adaptive fusion module and cross-modal attention module to fuse features from two modalities. Next, the progressive heterogeneous fusion module aggregates shallow features and semantic information. Then the fused features are integrated into decoders to get the fused images and segmentation results. Finally, the fused images are injected into SegNeXt to align the logits maps between teacher and student models before the backward.}
    \label{overflow}
\end{figure*}

\section{Proposed Method}
\label{sec3}
Firstly, we introduce the two training stages in Section \ref{two_training_stages}. Next, we put forward an infrared-visible image fusion and semantic segmentation framework to train both tasks simultaneously. Finally, we present the loss functions and the weight factors to balance the task-specific losses.

\subsection{Overall Framework}
\label{two_training_stages}
Unlike existing task-driven methods, we introduce knowledge distillation into the unified framework training procedure of fusion and segmentation, in which the logits maps make labels soft serving as a good regularization to the student network\cite{zhou2021wsl}. Despite one-hot label supervision, the student model also imitates the output of the teacher network to learn a latent representation of the segmentation task. Infrared image $I_{ir}\in \mathbb{R}^{H\times W\times 1}$ and visible image $I_{vi}\in \mathbb{R}^{H\times W\times 3}$ are defined as network inputs. We utilize the pre-trained encoder as the initialization and guidance for the second-stage multi-task training. This self-supervised pre-training strategy\cite{bachmann2022multimae} masks out many patches in the input image and predicts these missing regions, which motivates the network to study useful representations. In the first stage, the elaborated network reconstructs the original images. The reconstructed images are defined as $\hat{I}_{ir}$ and $\hat{I}_{vi} \in \mathbb{R}^{H\times W\times 1}$. In stage II, two task-specific decoders are applied to accomplish image fusion and semantic segmentation parallelly. We will demonstrate the proposed modules and training strategy in the following content.
\begin{figure}[h]
    \centering
    \includegraphics[width=0.9\linewidth]{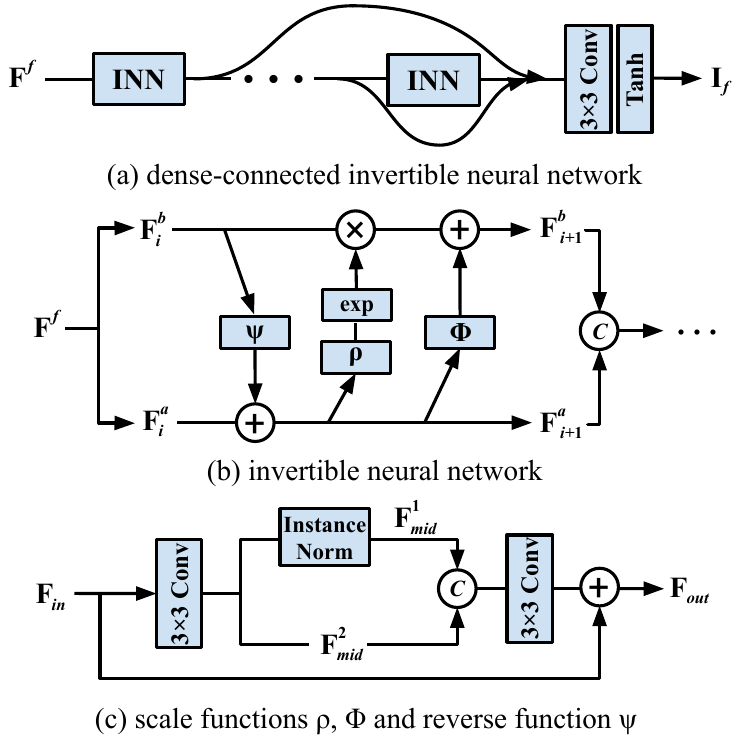}
    \caption{Architecture of proposed fusion decoder. Dense connections link the invertible units. Scale and reverse functions are implemented based on CNNs and instance norm.}
    \label{INN}
\end{figure}
\subsection{Feature extraction and fusion module}
\subsubsection{Shallow Feature Extraction}
Infrared image $I_{ir}\in \mathbb{R}^{H\times W\times 1}$ and visible image $I_{vi}\in \mathbb{R}^{H\times W\times 3}$ are inverted into the feature space with the same dimension $\mathbb{R}^{B\times C \times H\times W}$, in which $B$ is the batch size, $C$ stands for the current numbers of channels 64, $H$ and $W$ are the height and width of an image. Shallow features can be written as:
\begin{equation}
    \mathcal{F}_{vi}^{2} = SFE(I_{vi}), 
    \mathcal{F}_{ir}^{2} = SFE(I_{ir}), 
\end{equation}
in which $SFE(\cdot)$, the shallow feature extraction module is a preparation for the subsequent fusion from respective modalities as illustrated in Fig. \hyperref[overflow]{2(a)}. $\mathcal{F}_{vi}^{1}$ and $\mathcal{F}_{ir}^{1}$ are intermediate outputs of the $SFE$ procedure. $\mathcal{F}_{vi}^{2}$ and $\mathcal{F}_{ir}^{2}$ are the inputs of ResNet\cite{he2016deep} block and the dynamic extraction module (DEM) with a gate mechanism\cite{liang2023explicit} aims to rectify multi-scale deep features flexibly. The structure of $DEM$ is shown in Fig. \hyperref[overflow]{2(e)}. Thus we can get $\mathcal{F}_{vi}^{i}$ and $\mathcal{F}_{ir}^{i}$:
\begin{equation}
\begin{split}
    \mathcal{F}_{vi}^{i} = ResBlock(DEM(\mathcal{F}_{vi}^{i-1})), \\
    \mathcal{F}_{ir}^{i} = ResBlock(DEM(\mathcal{F}_{ir}^{i-1})),
\end{split}
\end{equation}
which are the deep semantic features extracted from the backbone network, and $i = 3, 4, 5, 6$. The specific fusion module for individual tasks will be discussed below.
\subsubsection{Cross-Modal Feature Fusion}On the one hand, we adopt the adaptive fusion strategy as shown in Fig. \hyperref[overflow]{2(b)} to obtain the shallow fusion features $\mathcal{F}^{i}$ defined as follows:
\begin{equation}
    \mathcal{F}^{i} = AFM(\mathcal{F}_{vi}^{i}, \mathcal{F}_{ir}^{i}), i = 1, 2.
\end{equation}
On the other hand, we utilize spatial and channel attention to get abundant cross-modal semantic information as demonstrated in Fig. \hyperref[overflow]{2(c)}. The deep fusion features $\mathcal{F}^{i}$ are defined as:
\begin{equation}
    \mathcal{F}^{i} = CAM(\mathcal{F}_{vi}^{i}, \mathcal{F}_{ir}^{i}), i = 3, 4, 5, 6.
\end{equation}
\label{shallow_deep_features}
We suppose that the shallow features for image fusion and the deep features for semantic segmentation are distinct and oriented for each task. The viewpoint that sharing the same encoder to seek consistency in feature space may hinder the performance of two tasks. To provide richer scene representation for semantic-aware image fusion, we emphasize that feature-level fusion tends to refine deep features with different resolutions, and image-level fusion involves the fusion of heterogeneous features, i.e., shallow features and semantic features.
\subsubsection{Progressive Heterogeneous Fusion} To provide better scene fidelity for image fusion, semantic information and shallow features are fused progressively by fully leveraging the heterogeneous features. The fused features $\mathcal{F}^{f}$ can be written as:
\begin{equation}
    \label{PHF_eq}
    \mathcal{F}^{f} = PHF(\mathcal{F}^{1}, PHF(\mathcal{F}^{2}, \mathcal{F}^{3})).
\end{equation}
The detailed structure is illustrated in Fig. \hyperref[overflow]{2(d)}. Then, visually perceptible fused images $I_{f} \in \mathbb{R}^{H\times W\times 1}$ are the generations from the fusion decoder.
\begin{figure}[h]
    \centering
    \includegraphics[width=0.9\linewidth]{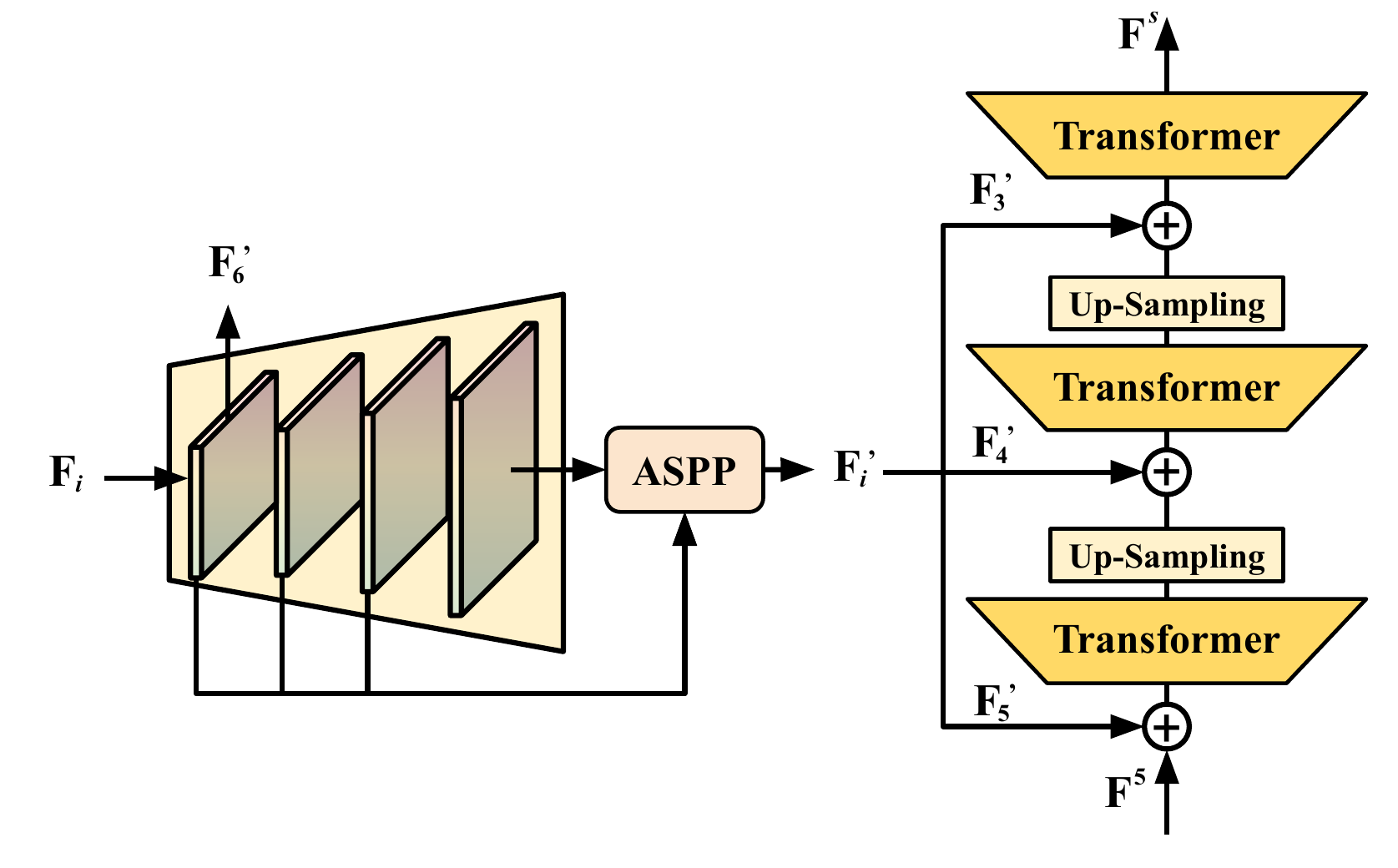}
    \caption{Architecture of the proposed segmentation decoder. Transformer blocks model long-range dependencies and aggregate fine-grained features via three stages in total. }
    \label{MST}
\end{figure}
\subsection{Fusion decoder}Recent studies \cite{zhou2022pan} indicated that the forward propagation of the CNN-based fusion network often suffers from information loss, especially high-frequency information. Intending to reduce the information loss as much as possible, we employed a dense-connected invertible neural network as the fusion decoder to get expected fused images with rich texture details (as shown in Fig. \ref{INN}). The fused features $\mathcal{F}^{f}$ are first evenly split into two components along the channel dimension: $\mathcal{F}^{a}_{i}$ and $\mathcal{F}^{b}_{i}$, and the intermediate outputs are calculated as:
\begin{equation}
    \label{INN_1}
    \mathcal{F}^{a}_{i+1} = \mathcal{F}^{a}_{i} + \psi(\mathcal{F}^{b}_{i}),
\end{equation}
\begin{equation}
    \label{exp}
    \mathcal{F}^{b}_{i+1} = \mathcal{F}^{b}_{i} \otimes exp(\rho(\mathcal{F}^{a}_{i+1})) + \phi(\mathcal{F}^{a}_{i+1}).
\end{equation}

In Eq. (\ref{INN_1}) and (\ref{exp}), $exp(\cdot)$ stands for the exponential function, and $\rho(\cdot)$ and $\phi(\cdot)$ are scale functions. $\psi(\cdot)$ denotes a reverse function. Notably, these functions are designed by neural network operations. $\mathcal{F}_{in}$ stands for the input of scale and reverse functions. And the output $\mathcal{F}_{out}$ is obtained through the process shown in Fig. \hyperref[INN]{3(c)}. Each later invertible unit is concatenated with the former output to reconstruct fused images via the residual channel attention operation, thereby producing the grayscale fused image $I_{f} \in \mathbb{R}^{H\times W\times 1}$ through a 3×3 convolutional layer followed by the Tanh activation function. We leverage the same YCbCr color space conversion operation as \cite{zhao2023cddfuse} to obtain the color image with the size of $H\times W\times 3$.

\subsection{Transformer-based segmentation decoder}After fully extracting cross-modal deep semantic features, considering the significance of establishing global context dependencies in the semantic segmentation task, a multi-stage transformer (MST) decoder is designed to model long-range spatial relationships, the refined semantic features $\mathcal{F}_{i}^{’}$ are obtained by:
\begin{equation}
    \label{MST_eq}
    \mathcal{F}_{i}^{’} = ASPP(FPN(\mathcal{F}^{i})).
\end{equation}
As shown in Fig. \ref{MST}, $FPN(\cdot )$ is the pixel decoder utilized in \cite{cheng2021per}, \cite{lin2017feature} to refine deep semantic features while maintaining the original feature size. And $ASPP(\cdot )$ is designed based on dilated convolution to increase the receptive field and accordingly improve multi-scale feature representation capability. After that, the inverted pyramid transformer decoder $MST(\cdot )$ aggregates multi-scale refined features as follows:
\begin{equation}
    \label{MST_eq2}
    \mathcal{F}^{s} = MST(\mathcal{F}^{5}, [\mathcal{F}_{3}^{’}\uparrow, \mathcal{F}_{4}^{’}\uparrow, \mathcal{F}_{5}^{’}\uparrow]),
\end{equation}
where $[\cdot]$ represents the concatenation at the channel dimension and $\uparrow$ is the upsample operation. The multi-head self-attention (MSA) and feed-forward network (FFN) inside the inverted pyramid transformer block are employed the same as the transformer in \cite{zhang2024semantic}. Then we obtain the final prediction maps $I_{p} \in \mathbb{R}^{B\times C\times H\times W}$ through a simple segmentation head, in which $C$ represents the number of classes.

\begin{algorithm}[t]
\caption{Two-Stage Network Training}
\label{alg}
\textbf{Require:} Pre-trained $SegNeXt$ model, randomly initialized $SFE$, $AFM$, $PHF$, $Backbone$, $CAM$, $De_{rec}$, $De_{fus}$, and $De_{seg}$ \\
\textbf{Input:} Infrared-visible image pairs $\{I_{vi}, I_{ir}\}$ and segmentation labels $I_l$ \\
\textbf{Output:} $SFE$, $AFM$, $PHF$, $Backbone$, $CAM$, $De_{fus}$, and $De_{seg}$ 
\begin{algorithmic}[1]
\STATE \textbf{First stage:}
\WHILE{not converged}
    \STATE Generate $\{\tilde{I}_{vi}, \tilde{I}_{ir}\}$ by adding 
 masking strategy
    \STATE Obtain features $\mathcal{F}^{f}$ by Eq. (\ref{PHF_eq})
    \STATE Reconstructed images $\hat{I}_{vi}, \hat{I}_{ir}$ = $De_{rec}(\mathcal{F}^{f})$
    \STATE Calculate $\mathcal{L}_{total}^{I}$ with $\{I_{vi}, I_{ir}\}$ by Eq. (\ref{totalI_eq})
    \STATE Update $SFE$, $AFM$, $PHF$, $Backbone$, $CAM$, and $De_{rec}$ by $\mathcal{L}_{total}^{I}$
\ENDWHILE

\STATE \textbf{Second stage:}
\STATE Load $SFE$, $AFM$, $PHF$, $Backbone$, and $CAM$
\WHILE{not converged}
    \STATE Generate features $\mathcal{F}^{f}$ by Eq. (\ref{PHF_eq})
    \STATE Fused images $I_{f} = De_{fus}(\mathcal{F}^{f})$
    \STATE Teacher logits maps $L_{h, w}^{t}$ = $SegNeXt$ ($I_{f}$)
    \STATE Obtain student logits maps $L_{h, w}^{s}$ and predicted maps $I_{p}$ from $De_{seg}$
    \STATE Calculate $\mathcal{L}_{total}^{II}$ with $\{I_{vi}, I_{ir}\}$ and $I_l$ by Eq. (\ref{15})
    \STATE Update $SFE$, $AFM$, $PHF$, $Backbone$, $CAM$, $De_{fus}$, and $De_{seg}$ by $\mathcal{L}_{total}^{II}$
\ENDWHILE
\end{algorithmic}
\end{algorithm}

\subsection{Network Training}
\label{sec_loss}
Considering the specific characteristics of infrared-visible image fusion and semantic segmentation, we design task-specific loss functions to guide our network for the implementation of two vision tasks. What's more, the dynamic weight factor has been introduced as a common technique for multi-task training. The proposed training strategy is summarized in Algorithm \ref{alg}. 
Now, we describe the detailed fusion loss, the semantic segmentation loss, and the balancing factor in sequence.
\begin{figure}[h]
    \centering
    \includegraphics[width=\linewidth]{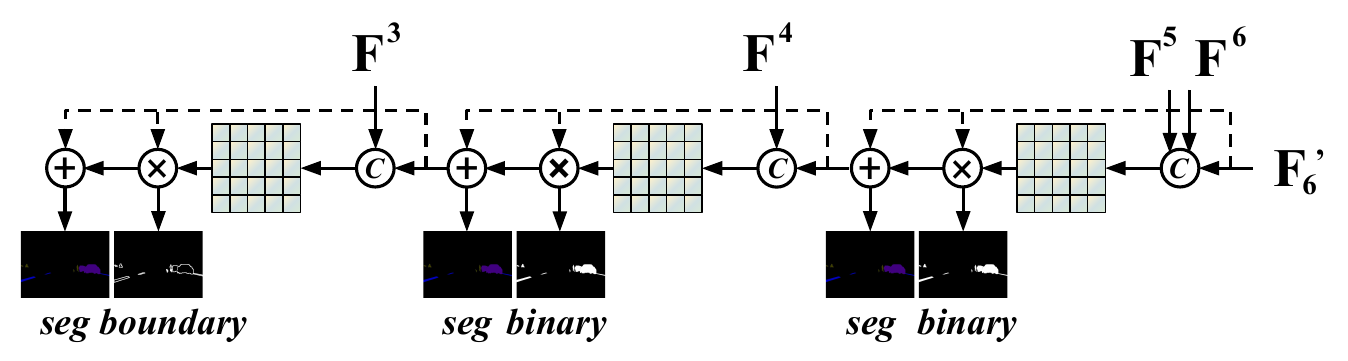}
    \caption{Architecture of the multilabel supervision part.  Auxiliary tasks have close relationships with the prediction of segmentation maps, which are used for reasoning feature-level fusion from two modalities.}
    \label{aux_pic}
\end{figure}
\subsubsection{Stage I}The total loss consists of reconstruction loss functions concerning visible and infrared images, respectively. Note that $\mathcal{L}_{vi}$ and $\mathcal{L}_{ir}$ only calculate the loss about the masked regions between reconstructed and original images. The total loss is listed as follows:
\begin{equation}
\label{totalI_eq}
\mathcal{L}_{total}^{I}= \mathcal{L}_{vi}+\mathcal{L}_{ir},
\end{equation}
where $\mathcal{L}_{vi}$ and $\mathcal{L}_{ir}$ are based on L1-norm with the following formulation, for visible images:
\begin{equation}
\label{rec}
\mathcal{L}_{vi} = \alpha_1 \mathcal{L}_{intensity}^{I} + \alpha_2 \mathcal{L}_{grad}^{I},
\end{equation}
\begin{equation}
\label{int1}
  \mathcal{L}_{intensity}^{I}=\frac{1}{HW}||\hat{I}_{vi} \otimes mask - I_{vi} \otimes mask||_1.
\end{equation}
In Eq. (\ref{rec}), $\alpha_1$ and $\alpha_2$ are both set as 5 according to our experience. In Eq. (\ref{int1}), $I_{vi}$ stands for the visible source image and $\hat{I}_{vi}$ corresponds to the reconstructed visible image.
\begin{equation}
\label{grad1}
  \mathcal{L}_{grad}^{I}=\frac{1}{HW}||\nabla (\hat{I}_{vi}\otimes mask) -\nabla (I_{vi}\otimes mask)||_1.
\end{equation}

The $mask$ defined by Eq. (\ref{int1}) and Eq. (\ref{grad1}) is the matrix calculated from the mask regions. We introduce the $mask$ into the reconstruction loss to focus on the masked regions as much as possible instead of taking an interest in the visible regions.

\subsubsection{Stage II}The total loss in the second stage consists of two parts: fusion loss and segmentation loss, as indicated by the following equation:
\begin{equation}
\label{15}
\mathcal{L}_{total}^{II}=\lambda_{f}\cdot\underbrace{(\mathcal{L}_{intensity}^{II}+\mathcal{L}_{grad}^{II})}_{Fusion \enspace loss}+\lambda_{s}\cdot\underbrace{(\mathcal{L}_{seg}+\mathcal{L}_{kd}+\mathcal{L}_{aux})}_{Segmentation\enspace loss}.
\end{equation}

For image fusion, we introduce intensity loss and gradient loss to constrain the pixel intensity and detail texture. 
\begin{equation}
  \mathcal{L}_{intensity}^{II}=\frac{1}{HW}||I_{f}-\max{(\left|I_{ir}\right|,\left|I_{vi}\right|)}||_1,
\end{equation}
\begin{equation}
  \mathcal{L}_{grad}^{II}=\frac{1}{HW}||\left|\nabla I_{f}\right|-\max{(\left|\nabla I_{ir}\right|,\left|\nabla I_{vi}\right|)}||_1,
\end{equation}
where $I_{f}$ represents for the fused image, $I_{vi}$ and $I_{ir}$ are paired source images and $\nabla (\cdot)$ is sobel gradient operator.
For semantic segmentation, we employ the OHEMCELoss\cite{shrivastava2016training} to calculate the semantic loss $L_{seg}$ for the final output and the semantic constraints in the multilabel supervision (as displayed in Fig. \ref{aux_pic}), to mitigate the challenges posed by hard samples.
\begin{equation}
    \label{aux}
    \mathcal{L}_{aux} = \mathcal{L}_{boundary} + \mathcal{L}_{binary} + \mathcal{L}_{seg}.
\end{equation}

In Eq. (\ref{aux}), the boundary semantic loss $\mathcal{L}_{boundary}$ is designed to enhance the edge details of objects. And the binary semantic loss $\mathcal{L}_{binary}$ has the strength of reducing the interference of cluttered backgrounds. Cross-entropy loss is chosen to calculate the boundary semantic loss. Due to the class imbalance problem between the background and objects, the weighted cross-entropy loss is utilized to calculate the binary semantic loss. $\mathcal{L}_{binary}$ is listed as follows:
\begin{equation}
    \begin{split}
    \mathcal{L}_{binary} = \sum_{i=1}^{N} w \cdot ({l}_{x,y}^{i} \times log({p}_{x,y}^{i}) \\
    +(1-{l}_{x,y}^{i}) \times log(1-{p}_{x,y}^{i})),
    \end{split}
\end{equation}
in which ${l}_{x,y}^{i}$ denotes the pixel of ground-truth label, ${p}_{x,y}^{i}$ corresponds to the predicted map's pixel and $N$ is the total number of categories.
To accomplish the alignment of the results between teacher and student networks\cite{zhou2021rethinking},\cite{hao2024primkd},\cite{qiu2024make}, knowledge distillation loss is designed to constrain their logits maps as similar as possible, which is formulated as follows:
\begin{equation}
    \mathcal{L}_{kd} = \sum_{h=1}^{H} \sum_{w=1}^{W} KL(\rho(\frac{L_{h,w}^{s}}{T})||\rho(\frac{L_{h,w}^{t}}{T})),
\end{equation}
where the temperature $T$ is empirically set as 4, $KL(\cdot)$ stands for calculating the Kullback–Leibler (KL) divergence, $L_{h,w}^{t}$ is the logits map of teacher network and $L_{h,w}^{s}$ is the logits map of student network.

\subsubsection{Dynamic weight factor}In the process of collaborative learning between image fusion and semantic segmentation, the magnitudes of losses for both tasks tend to vary, which inspired us to find re-weighting rules to balance tasks better. We denote a specific-task loss set $\mathcal{L}(\theta) = \{\mathcal{L}_{f}(\theta), \mathcal{L}_{s}(\theta)\}$, where $\theta$ represents the network parameters, $\mathcal{L}_{f}$ is the \textbf{\textit{Fusion loss}} and $\mathcal{L}_{s}$ is the \textbf{\textit{Segmentation loss}} as shown in Eq.(\ref{15}). Firstly, we introduce $\alpha$-fairness\cite{ban2024fair} allocation principle into the task losses, the losses $\{\mathcal{L}_{f}, \mathcal{L}_{s}\}$ are replaced with following formulations:
\begin{equation}
    \{\frac{\mathcal{L}_{f}^{1 - \alpha}}{1 - \alpha}, \frac{\mathcal{L}_{s}^{1 - \alpha}}{1 - \alpha}\},
\end{equation}
where $\alpha\in(-\infty, 1)$. Max-min fairness tends to give the {\itshape ``less-fortune''} task a smaller gradient magnitude in practice\cite{ban2024fair}. Since the segmentation task poses greater optimization challenges, the $\alpha$ in $\mathcal{L}(\theta)$ is set as 0.8. Then, the Dynamic Weight Average (DWA)\cite{liu2019end} strategy is employed to dynamically reweight specific task losses $\mathcal{L}_{f}$ and $\mathcal{L}_{s}$. These settings will also be discussed in Section \ref{sec4_DWA}. 
\begin{equation}
    \lambda_k(t+1) = \frac{K\exp(w_k(t)/T)}{\sum_{i=1}^{K} \exp(w_i(t)/T)}, w_k(t) = \frac{\mathcal{L}_k(\theta_t)}{\mathcal{L}_k(\theta_{t - 1})},
\end{equation}
where $T$ is the temperature parameter set as 500. $K$ denotes the total number of tasks, and $k$ refers to the specific task. And $t$ in $\lambda_k(t+1)$ denotes the iteration in the current training epoch. $w_k(t)$ is concluded by calculating the task-specific loss $\mathcal{L}_k(\theta_t)$ in the {\itshape t}-th iteration and $\mathcal{L}_k(\theta_{t - 1})$ from the last iteration.

\section{Experimental Results}
\label{sec4}
In this section, to better display the efficacy of our proposed simple yet efficient joint learning paradigm, we implement the experiments by answering the following questions:
\begin{itemize}
\item $\mathbf{RQ1}$: Which datasets and detailed experimental settings are used for training and evaluation?
\item $\mathbf{RQ2}$: Have we succeeded in promising results in objective and subjective comparisons?
\item $\mathbf{RQ3}$: How do masking techniques, network architectures, and training strategies influence fusion and segmentation joint learning?
\item $\mathbf{RQ4}$: Is the proposed approach reasonable and effective?
\end{itemize}
\subsection{Experiment Configurations ($\mathbf{RQ1}$)}

\begin{figure*}[h]
    \centering
    \includegraphics[width=\linewidth]{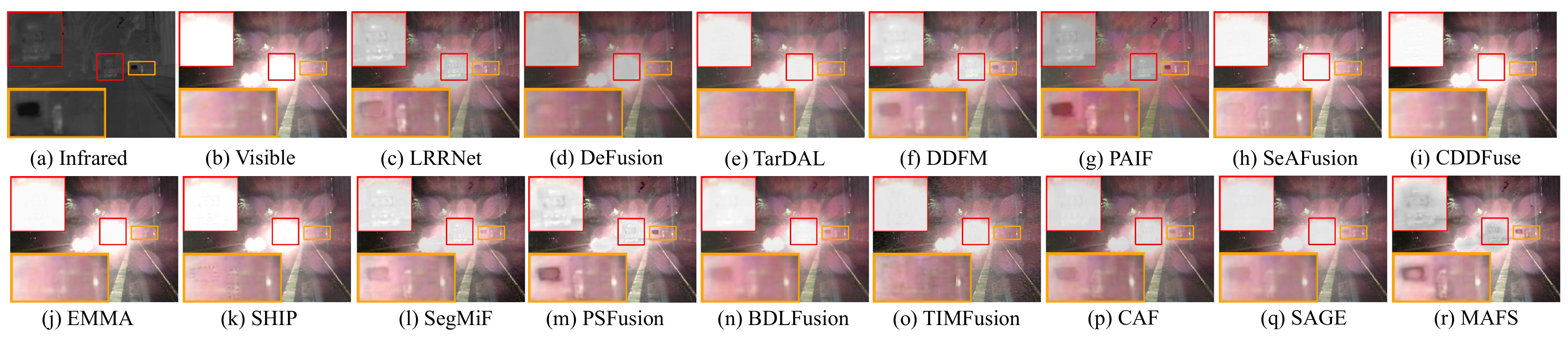}
    \caption{Qualitative fusion results on MFNet \cite{ha2017mfnet} dataset. The regions enclosed by the red and orange rectangles are magnified to highlight the truck in intense lighting conditions and the distant pedestrian near the roadside, respectively.}
    \label{MFNet}
\end{figure*}

\begin{figure*}[h]
    \centering
    \includegraphics[width=\linewidth]{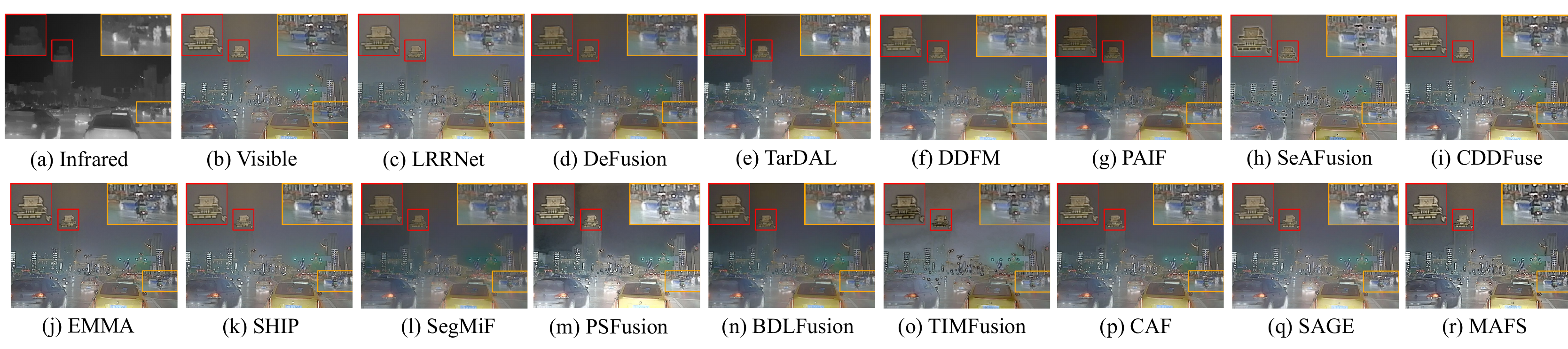}
    \caption{Qualitative fusion results on FMB\cite{liu2023multi} dataset. The regions enclosed by the red and orange rectangles are magnified to describe a distant building at night and the taxi, the motorcyclist, and several pedestrians, respectively.}
    \label{FMB}
\end{figure*}

\begin{figure*}[h]
    \centering
    \includegraphics[width=\linewidth]{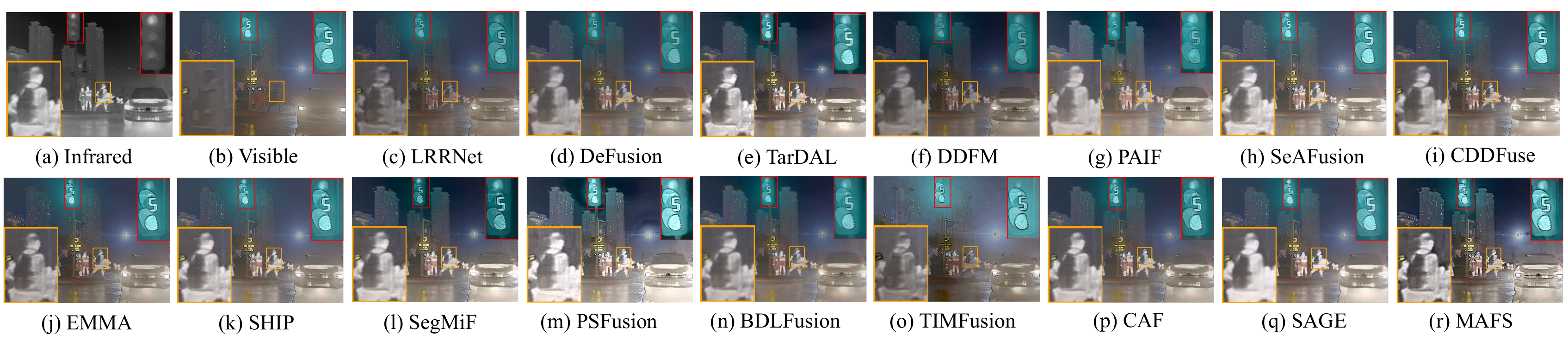}
    \caption{Qualitative fusion results on M$^{3}$FD \cite{liu2022target} dataset. The regions enclosed by the red and orange rectangles are magnified to show the traffic light at the intersection and the person wearing a backpack, respectively.}
    \label{M3FD}
\end{figure*}

\subsubsection{Datasets}We trained the network on the MFNet\cite{ha2017mfnet}, PST900\cite{shivakumar2020pst900} and FMB\cite{liu2023multi} datasets. The MFNet has nine classes with 1,569 pairs of RGB and thermal images, covering 820 pairs of daytime scenes and 749 pairs of nighttime scenes. The resolution is 640$\times$480 pixels. The PST900 dataset has five classes with 894 pairs of RGB and thermal images. And its resolution is 640$\times$1280 pixels. The FMB dataset encompasses 1500 image pairs with a resolution of 600$\times$800 and has fifteen categories for semantic segmentation. The dataset splits of MFNet, PST900 and FMB are the same as the original paper. Furthermore, we select 420 M$^{3}$FD\cite{liu2022target} pairs at equal intervals as the fusion test set, a mainstream benchmark in the infrared-visible image fusion field. 
\subsubsection{Training Configuration}We conducted the PyTorch 2.3.1, CUDA 11.8 to implement MAFS. A machine equipped with two NVIDIA GeForce 4090 graphics cards was utilized for the whole experimental process.
We used general data augmentation ways, i.e., randomly cropping images into a size of 256$\times$256, random flipping, and 90-degree rotation in two training stages. The batch size was set as 4. We chose ResNet-152\cite{he2016deep} as the backbone. As an advanced unimodal segmentation framework, SegNeXt\cite{guo2022segnext} was pre-trained on MFNet, PST900, and FMB, respectively, to adapt to the corresponding tasks. The first stage has 100 epochs, and then the second-stage model was trained for 400 epochs to achieve minimal training losses. Both training stages utilized the SGD optimizer with an initial learning rate of 1e-3 and an exponential weight decay.
\subsubsection{State-of-the-art Methods and evaluation metrics}We selected fifteen state-of-the-art methods to assess fusion results. They are SeAFusion\cite{tang2022image}, DDFM\cite{zhao2023ddfm}, PAIF\cite{liu2023paif}, DeFusion\cite{liang2022fusion}, TarDAL\cite{liu2022target}, LRRNet\cite{li2023lrrnet}, CDDFuse\cite{zhao2023cddfuse}, EMMA\cite{zhao2024equivariant}, SHIP\cite{zheng2024probing}, SegMiF\cite{liu2023multi}, PSFusion\cite{tang2023rethinking}, BDLFusion\cite{liu2023bilevel}, TIMFusion\cite{TIMFusion}, CAF\cite{CAF}, and SAGE\cite{wu2025every}. Make sure fair comparisons, 
 followed by CDDFuse\cite{zhao2023cddfuse} and EMMA\cite{zhao2024equivariant}, entropy (EN), standard deviation (SD), spatial frequency (SF), visual information fidelity (VIF), gradient-based similarity measurement (Qabf), and average gradient (AG) are evaluation metrics for quantitative comparisons on infrared-visible image fusion test sets: MFNet, FMB and M$^{3}$FD. For the semantic segmentation task, we compare MAFS with SeAFusion\cite{tang2022image}, SegMiF\cite{liu2023multi}, PAIF\cite{liu2023paif}, PSFusion\cite{tang2023rethinking}, MFNet\cite{ha2017mfnet}, EGFNet\cite{dong2023egfnet}, RTFNet\cite{sun2019rtfnet}, LASNet\cite{li2022rgb} and GMNet\cite{zhou2021gmnet} on the MFNet dataset. In addition, we also compared quantitative results with ACNet\cite{hu2019acnet}, SA-Gate\cite{chen2020bi}, MFNet\cite{ha2017mfnet}, PSTNet\cite{shivakumar2020pst900}, MFFENet\cite{zhou2021mffenet}, EGFNet\cite{dong2023egfnet}, MTANet\cite{zhou2022mtanet}, FDCNet\cite{zhao2022feature} and MMSMCNet\cite{zhou2023mmsmcnet} on the PST900 dataset. Extension experiments on the FMB dataset are compared with previous collaborative learning methods, i.e., SeAFusion\cite{tang2022image}, PSFusion\cite{tang2023rethinking}, and SegMiF\cite{liu2023multi}. The mean intersection over union (mIoU), a widely used evaluation metric, is employed to quantify scene-parsing performance.

\begin{table}[!ht]
    \centering
    \caption{The average EN, SD, SF, VIF, Qabf and AG scores on the MFNet\cite{ha2017mfnet} dataset for infrared-visible fusion task. (\textcolor{red}{\textbf{Red}}: best; \textcolor{blue}{\textbf{Blue}}: second best.)}
    \label{MFNet_tab}
    \resizebox{\linewidth}{!}{
    \begin{tabular}{ c | c  c  c  c  c  c}
    \toprule
        \multicolumn{1}{c|}{\multirow{2}{*}{\textbf{Method}}} & \multicolumn{6}{c}{\textbf{Dataset: MFNet Fusion Dataset}} \\ 
        ~ & EN &  SD &  SF &  VIF &  Qabf &  AG \\ \midrule
        LRRNet\cite{li2023lrrnet} & 6.3787 & 32.3543 & 9.1470 & 0.7167 & 0.5278 & 4.1665 \\ 
        DeFusion\cite{liang2022fusion} & 6.0393 & 24.0992 & 6.2040 & 0.7201 & 0.3944 & 2.9780 \\ 
        TarDAL\cite{liu2022target} & 6.3417 & 31.0926 & 9.5281 & 0.6799 & 0.4431 & 4.0806 \\ 
        DDFM\cite{zhao2023ddfm} & 6.4825 & 30.7697 & 8.2823 & 0.8544 & 0.5797 & 3.9022  \\ 
        PAIF\cite{liu2023paif} & 6.0660 & 21.6480 & 7.5953 & 0.7306 & 0.4114 & 3.5587 \\ 
        SeAFusion\cite{tang2022image} & 6.4574 & 34.4890 & 11.5296 & 0.8090 & 0.6402 & 5.3914 \\ 
        CDDFuse\cite{zhao2023cddfuse} & 6.4951 & 38.0910 & 11.6775 & 0.9513 & 0.6602 & 5.1424 \\ 
        EMMA\cite{zhao2024equivariant} & 6.6123 & 40.3196 & 12.2276 & 0.9042 & 0.6420 & 5.6596 \\ 
        SHIP\cite{zheng2024probing} & 6.4739 & 34.7598 & 12.1840 & 0.8908 & 0.6525 & 5.6713 \\
        SegMiF\cite{liu2023multi} & 6.6584 & 39.6719 & 11.4429 & 0.8153 & 0.6258 & 5.436 \\
        PSFusion\cite{tang2023rethinking} & 6.9287 & \textcolor{blue}{\textbf{42.5341}} & \textcolor{blue}{\textbf{14.504}} & \textcolor{blue}{\textbf{1.0296}} & \textcolor{blue}{\textbf{0.6771}} & \textcolor{blue}{\textbf{6.7932}} \\
        BDLFusion\cite{liu2023bilevel} & 6.2758 & 28.8315 & 7.433 & 0.8437 & 0.5193 & 3.567 \\
        TIMFusion\cite{TIMFusion} & \textcolor{red}{\textbf{7.0845}} & 42.2399 & 11.6902 & 0.4593 & 0.3967 & 5.922 \\
        CAF\cite{CAF} & 6.3844 & 30.9068 & 12.322 & 0.6533 & 0.4803 & 4.8894 \\
        SAGE\cite{wu2025every} & 6.2922 & 36.1515 & 10.9054 & 0.8056 & 0.6302 & 4.92 \\
        MAFS & \textcolor{blue}{\textbf{7.0033}} & \textcolor{red}{\textbf{46.6232}} & \textcolor{red}{\textbf{15.3121}} & \textcolor{red}{\textbf{1.1437}} & \textcolor{red}{\textbf{0.7102}} & \textcolor{red}{\textbf{7.1457}} \\ \bottomrule
    \end{tabular}
    }
\end{table}

\begin{table}[!ht]
    \centering
    \caption{The average EN, SD, SF, VIF, Qabf and AG scores on the FMB\cite{liu2023multi} dataset for infrared-visible fusion task. (\textcolor{red}{\textbf{Red}}: best; \textcolor{blue}{\textbf{Blue}}: second best.)}
    \label{FMB_tab}
    \resizebox{\linewidth}{!}{
    \begin{tabular}{ c | c  c  c  c  c  c}
    \toprule
        \multicolumn{1}{c|}{\multirow{2}{*}{\textbf{Method}}} & \multicolumn{6}{c}{\textbf{Dataset: FMB Fusion Dataset}} \\ 
        ~ & EN &  SD &  SF &  VIF &  Qabf &  AG \\ \midrule
        LRRNet\cite{li2023lrrnet} & 6.2807 & 26.2629 & 10.1829 & 0.6037 & 0.5509 & 4.3569 \\ 
        DeFusion\cite{liang2022fusion} & 6.3885 & 27.9476 & 7.1399 & 0.5908 & 0.3716 & 3.1553 \\ 
        TarDal\cite{liu2022target} & 6.9588 & 40.2792 & 11.5769 & 0.6567 & 0.4536 & 4.9225 \\ 
        DDFM\cite{zhao2023ddfm} & 6.4929 & 27.7969 & 7.8937 & 0.6307 & 0.4537 & 3.4057 \\ 
        PAIF\cite{liu2023paif} & 6.7143 & 34.8001 & 8.4258 & 0.6198 & 0.409 & 3.769 \\ 
        SeAFusion\cite{tang2022image} & 6.5194 & 28.7364 & 13.9664 & 0.1559 & 0.6158 & 6.0595 \\ 
        CDDFuse\cite{zhao2023cddfuse} & 6.7761 & 37.0318 & 14.5628 & 0.8784 & \textcolor{blue}{\textbf{0.6787}} & 6.0726 \\ 
        EMMA\cite{zhao2024equivariant} & 6.7729 & 36.7961 & 14.9961 & 0.8329 & 0.6481 & 6.6106 \\ 
        SHIP\cite{zheng2024probing} & 6.6574 & 34.0897 & 14.5799 & \textcolor{blue}{\textbf{0.894}} & \textcolor{red}{\textbf{0.6968}} & 6.2832 \\ 
        SegMiF\cite{liu2023multi} & 6.5054 & 27.6731 & 7.9276 & 0.6106 & 0.4325 & 3.4682 \\
        PSFusion\cite{tang2023rethinking} & \textcolor{red}{\textbf{7.3088}} & \textcolor{red}{\textbf{48.927}} & \textcolor{blue}{\textbf{18.1556}} & 0.8481 & 0.627 & \textcolor{blue}{\textbf{7.9855}} \\
        BDLFusion\cite{liu2023bilevel} & 6.6984 & 29.7085 & 8.7806 & 0.622 & 0.449 & 4.3532 \\
        TIMFusion\cite{TIMFusion} & 6.4983 & 32.4909 & 12.9743 & 0.5199 & 0.5484 & 5.3377 \\
        CAF\cite{CAF} & 6.6958 & 32.2307 & 12.7604 & 0.5946 & 0.5131 & 5.354\\
        SAGE\cite{wu2025every} & 6.8287 & 37.5874 & 12.6436 & 0.7704 & 0.6372 & 5.3522 \\
        MAFS & \textcolor{blue}{\textbf{7.1309}} & \textcolor{blue}{\textbf{40.3626}} & \textcolor{red}{\textbf{20.1923}} & \textcolor{red}{\textbf{0.9609}} & 0.631 & \textcolor{red}{\textbf{9.1019}} \\ 
        \bottomrule
    \end{tabular}
    }
\end{table}

\subsection{Infrared-Visible Image Fusion ($\mathbf{RQ2}$)}
\label{4-2}

\subsubsection{Subjective Comparision}
As shown in Fig. \ref{MFNet}, due to the dramatic lighting changes in low-light environments, the halo effect created by headlights affects the appearance of objects around the vehicle presented in the visible image. DeFusion, TarDAL, SeAFusion, CDDFuse, EMMA, SHIP, and TIMFusion encounter the above issue. 
Considering the visual enhancement under extreme scenarios, PAIF, SegMiF, PSFusion, BDLFusion, CAF, and SAGE excel in stable performance and highlight the person at a distance. And BDLfusion, CAF, and SAGE show limited effectiveness in enhancing vehicles under strong light conditions. The proposed MAFS highlights the pedestrian and enhances the vehicle's texture and shape.
As illustrated in Fig. \ref{FMB}, MAFS enhances the detailed textures of the left building by observing the magnified area at the top of the building's structure. Notably, all the listed fused images highlight the people walking along the street on the right and the person driving on the motorbike. 
PSFusion achieves effective preservation of intensity and texture information, and also exhibits high-quality visual results. Intuitively, our proposed method produces the fused image with the best contrast and preserves abundant textures from the source visible image in the nighttime. In terms of the M$^{3}$FD dataset, it's a crowded intersection consisting of cars and pedestrians shown in Fig. \ref{M3FD}. TarDAL, SeAFusion, CDDFuse, EMMA, SHIP, PSFusion, CAF, and SAGE fully retain the brightness and structural features of the traffic light in the visible image. For detailed texture preservation, TarDAL, SeAFusion, CDDFuse, SHIP, SegMiF, BDLFusion, and CAF enable to enhance salient objects, maintaining consistency with the source images. MAFS enhances the contour of traffic lights. The number {\textit{``5''}} maintains high contrast and color saturation, while the highlighted pedestrian exhibits rich texture details and color saturation.

\subsubsection{Objective Comparision}
We show the quantitative results on six evaluation metrics. According to the MFNet dataset, MAFS achieves first place in five out of six metrics as listed in Table \ref{MFNet_tab}. And TIMFusion obtains the highest value of entropy. In addition, PSFusion attains the second-best ranking in five metrics, highlighting its outstanding fusion performance. 
In terms of the FMB dataset, MAFS balances low-light road scenarios and enhances edge details for some local regions so that fused images present the best value of SF, VIF, and AG in Table \ref{FMB_tab}. 
SHIP utilizing high-order interaction, highlights the salient thermal targets and preserves rich information, which is dominant for the subsequent application. PSFusion has the best EN and SD, which indicate the highest contrast and the richest details in fused images.
As for the M$^{3}$FD dataset, containing most of the road scenarios, MAFS surpasses other state-of-the-art methods on EN, SF, VIF, and AG. SegMiF exhibits high-quality visual results. PSFusion achieves abundant detail texture and sharp edges in the fused images.
We speculate that the semantic information contained by fused images benefits high-level tasks, e.g., semantic segmentation. This hypothesis will be verified in the next subsection.

\begin{table}[!ht]
    \centering
    \caption{The average EN, SD, SF, VIF, Qabf and AG scores on the M$^{3}$FD\cite{liu2022target} dataset for infrared-visible fusion task. (\textcolor{red}{\textbf{Red}}: best; \textcolor{blue}{\textbf{Blue}}: second best.)}
    \label{M3FD_tab}
    \resizebox{\linewidth}{!}{
    \begin{tabular}{c | c  c  c  c  c  c}
    \toprule
        \multicolumn{1}{c|}{\multirow{2}{*}{\textbf{Method}}} & \multicolumn{6}{c}{\textbf{Dataset: M$^{3}$FD Fusion Dataset}} \\ 
        ~ & EN &  SD &  SF &  VIF &  Qabf &  AG \\ \midrule
        LRRNet\cite{li2023lrrnet} & 6.4699 & 27.7628 & 11.1664 & 0.5572 & 0.5084 & 5.2133 \\ 
        DeFusion\cite{liang2022fusion} & 6.4964 & 27.7765 & 7.7281 & 0.5308 & 0.3340 & 3.7586 \\ 
        TarDAL\cite{liu2022target} & 6.9905 & 38.7646 & 12.9874 & 0.5742 & 0.4220 & 6.1445 \\ 
        DDFM\cite{zhao2023ddfm} & 6.6775 & 30.1044 & 9.3393 & 0.6168 & 0.4726 & 4.4978 \\ 
        PAIF\cite{liu2023paif} & 6.9826 & 37.1753 & 9.5188 & 0.5881 & 0.3966 & 4.6549 \\ 
        SeAFusion\cite{tang2022image} & 6.8866 & 36.0223 & 15.3075 & 0.6749 & 0.6066 & 7.4716 \\ 
        CDDFuse\cite{zhao2023cddfuse} & 6.9393 & 37.8472 & 15.4533 & 0.7684 & 0.6192 & 7.0522 \\ 
        EMMA\cite{zhao2024equivariant} & 6.9806 & 39.2056 & 15.8641 & 0.7487 & 0.6026 & 7.7150 \\ 
        SHIP\cite{zheng2024probing} & 6.8824 & 36.1726 & 15.9912 & \textcolor{blue}{\textbf{0.7997}} & \textcolor{blue}{\textbf{0.6489}} & 7.5437 \\
        SegMiF\cite{liu2023multi} & 7.0717 & 39.7735 & 14.7833 & 0.7688 & \textcolor{red}{\textbf{0.6491}} & 6.9532\\
        PSFusion\cite{tang2023rethinking} & \textcolor{blue}{\textbf{7.3685}} & \textcolor{red}{\textbf{49.4131}} & \textcolor{blue}{\textbf{21.6251}} & 0.7834 & 0.5775 & \textcolor{blue}{\textbf{10.0291}} \\
        BDLFusion\cite{liu2023bilevel} & 6.6984 & 29.7085 & 8.7806 & 0.622 & 0.449 & 4.3532 \\
        TIMFusion\cite{TIMFusion} & 6.7001 & 39.4562 & 13.659 & 0.5285 & 0.5074 & 6.1753 \\
        CAF\cite{CAF} & 6.7951 & 32.5505 & 13.4877 & 0.5447 & 0.4681 & 6.2004 \\
        SAGE\cite{wu2025every} & 6.8746 & 37.0387 & 13.348 & 0.6949 & 0.5875 & 6.1979 \\
        MAFS & \textcolor{red}{\textbf{7.3751}} & \textcolor{blue}{\textbf{47.3692}} & \textcolor{red}{\textbf{22.9149}} & \textcolor{red}{\textbf{0.918}} & 0.5965 & \textcolor{red}{\textbf{10.9709}} \\ \bottomrule
    \end{tabular}
    }
\end{table}

\begin{figure*}[h]
    \centering
    \includegraphics[width=\linewidth]{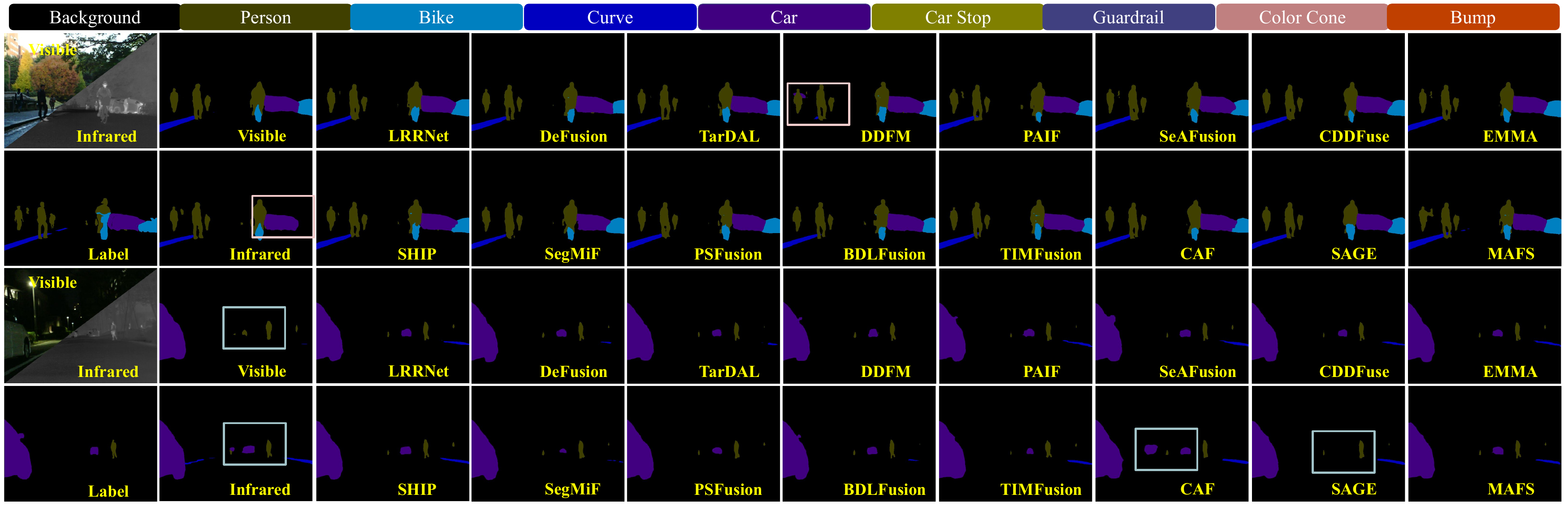}
    \caption{Visual comparisons with fifteen SOTA fusion methods in daytime and nighttime images of MFNet\cite{ha2017mfnet} dataset. Rectangular boxes on the segmentation results denote regions of misclassification or omission.}
    \label{MF_seg2}
\end{figure*}

\begin{table*}[h!]
\centering
\caption{Quantitative results (\%) of post-fusion segmentation strategy compared with fifteen SOTA fusion methods on the MFNet\cite{ha2017mfnet} dataset. The value 0.0 means that there are no true positives. (\textcolor{red}{\textbf{Red}}: best; \textcolor{blue}{\textbf{Blue}}: second best.)}
\label{MF_segtab2}
\resizebox{0.7\textwidth}{!}{
\begin{tabular}{c|cccccccc|c}
\toprule
\textbf{Method} & \textbf{Car} & \textbf{Person} & \textbf{Bike} & \textbf{Curve} & \textbf{Car Stop} & \textbf{Guardrail} & \textbf{Cone} & \textbf{Bump} & \textbf{mIoU} \\ \midrule
        Visible & 88.31 & 63.19 & 63.24 & 36.33 & 30.54 & 3.13 & 53.35 & 45.91 & 53.54 \\ 
        Infrared & 86.57 & 71.63 & 59.01 & 39.27 & 21.33 & 6.55 & 43.7 & 49.5 & 52.82 \\ \midrule
        LRRNet\cite{li2023lrrnet} & 88.85 & 72.62 & 63.76 & 43.7 & 30.92 & 3.89 & 53.36 & 46.76 & 55.78 \\ 
        DeFusion\cite{liang2022fusion} & 89.18 & 73.67 & 63.98 & 46.82 & 26.59 & 0.1 & 51.66 & 57.29 & 56.39 \\ 
        TarDAL\cite{liu2022target} & 88.47 & 71.09 & 64.23 & 42.81 & 31.26 & \textcolor{red}{\textbf{8.47}} & 51.7 & 49.88 & 56.22 \\ 
        DDFM\cite{zhao2023ddfm} & 88.84 & 73.07 & 64.86 & 44.23 & 29.24 & 0 & 53.26 & 52.26 & 55.99 \\ 
        PAIF\cite{liu2023paif} & \textcolor{red}{\textbf{89.5}} & 73.67 & 64.32 & 44.86 & 26.83 & 7.2 & 53.12 & \textcolor{blue}{\textbf{58.56}} & 57.36 \\
        SeAFusion\cite{tang2022image} & 88.88 & 71.97 & 65.34 & 46.56 & 29.84 & 1.59 & 52.22 & 50.07 & 56.07 \\ 
        CDDFuse\cite{zhao2023cddfuse} & 88.13 & 73.59 & 63.45 & 46.92 & 24.58 & 0 & 53.22 & 48.16 & 55.13 \\ 
        EMMA\cite{zhao2024equivariant} & \textcolor{blue}{\textbf{89.31}} & 73.42 & 64.81 & \textcolor{blue}{\textbf{46.97}} & \textcolor{blue}{\textbf{35.36}} & 0.38 & 51.41 & 50.79 & 56.77 \\ 
        SHIP\cite{zheng2024probing} & 88.6 & \textcolor{blue}{\textbf{73.87}} & 64.99 & 46.8 & 27.49 & 4.64 & \textcolor{red}{\textbf{54.4}} & 54.2 & 57.03 \\ 
        SegMiF\cite{liu2023multi} & 88.4 & 72.34 & 64.12 & 44.99 & 34.99 & 0 & 51.33 & \textcolor{red}{\textbf{60.57}} & 57.21 \\ 
        PSFusion\cite{tang2023rethinking} & 88.9 & 72.95 & \textcolor{blue}{\textbf{65.49}} & 46.67 & 29.42 & 5.91 & 53.3 & 55.54 & \textcolor{blue}{\textbf{57.38}} \\
        BDLFusion\cite{liu2023bilevel} & 89.16 & \textcolor{red}{\textbf{74.29}} & 63.81 & \textcolor{red}{\textbf{47.97}} & 25.37 & 0 & 52.58 & 57.12 & 56.50  \\ 
        TIMFusion\cite{TIMFusion} & 87.44 & 70.55 & 62.7 & 36.34 & 24.89 & 6.2 & 50.84 & 44.25 & 53.47 \\ 
        CAF\cite{CAF} & 87.04 & 71.97 & 64.71 & 44.37 & 32.24 & \textcolor{blue}{\textbf{7.88}} & 49.15 & 48.47 & 55.99 \\
        SAGE\cite{wu2025every} & 87.78 & 72.13 & 64.59 & 39.36 & 31.87 & 4.5 & 51.8 & 47.3 & 55.26 \\\midrule
        MAFS & 88.8 & 72.24 & \textcolor{red}{\textbf{65.52}} & 45.23 & \textcolor{red}{\textbf{35.55}} & 6.06 & \textcolor{blue}{\textbf{53.98}} & 51.32 & \textcolor{red}{\textbf{57.44}} \\ 
\bottomrule
\end{tabular}
}
\end{table*}

\begin{figure*}[h]
    \centering
    \includegraphics[width=\linewidth]{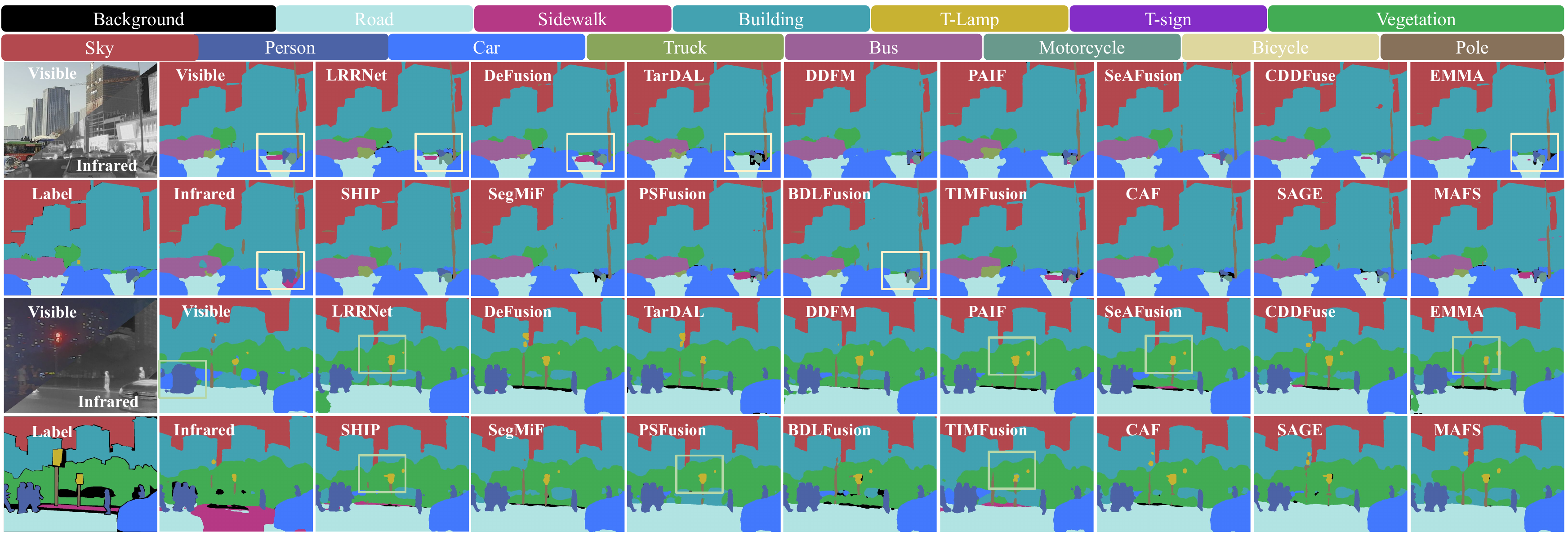}
    \caption{Visual comparisons with fifteen SOTA fusion methods in daytime and nighttime images of FMB\cite{liu2023multi} dataset. Rectangular boxes on the segmentation results denote regions of misclassification or omission.}
    \label{FMB_seg2}
\end{figure*}

\subsubsection{Adaptation for Semantic Segmentation}
We choose SegNeXt-Large\cite{guo2022segnext} as the segmentation backbone, which pays more attention to efficiency for semantic segmentation. As fair comparisons for the segmentation performance of state-of-the-art (SOTA) fusion methods, we follow their default settings to generate the required fused images of the MFNet\cite{ha2017mfnet} and FMB\cite{liu2023multi} datasets. For qualitative evaluation, as illustrated in Fig. \ref{MF_seg2}, due to the advantages of multimodal image fusion, the fused images effectively combine the complementary characteristics and attain high segmentation performance. In particular, DDFM encounters difficulties in segmenting the curve along the road. Other fusion methods demonstrate better semantic representation capabilities in daytime scenarios. Segmenting distant people and cars often remains a significant challenge at night, mainly due to their small scale and low visibility in complex scenarios. MAFS achieves more accurate segmentation of people and cars, benefiting from richer semantic information in the fused images. According to the FMB dataset, 
from comparisons between the third and fourth rows in Fig. \ref{FMB_seg2}, it is evident that MAFS produces finer boundaries and segments better on the Lamp category. For quantitative evaluation, the results of the post-fusion segmentation compared with state-of-the-art methods are listed in Table \ref{MF_segtab2} and \ref{FMB_segtab2}. As for the MFNet\cite{ha2017mfnet} dataset, the proposed MAFS achieves the best IoU of Bike, Car Stop, and Cone categories. PSFusion demonstrates the second-best mIoU, which maintains useful semantic information.
MAFS surpasses PSFusion 0.06\% in mIoU on the MFNet dataset. 
For the FMB\cite{liu2023multi} dataset, MAFS presents the highest Iou of the Truck category. SHIP excels in segmenting people in complex environments. PAIF enables segmenting the Truck accurately. BDLFusion, formulating the inner mutual optimization, obtains the second-best results and also displays satisfactory semantic performance.

\subsection{Infrared-Visible Semantic Segmentation ($\mathbf{RQ2}$)}
\label{4-3}
\subsubsection{Subjective Comparision}
\begin{figure*}[h]
    \centering
    \includegraphics[width=\linewidth]{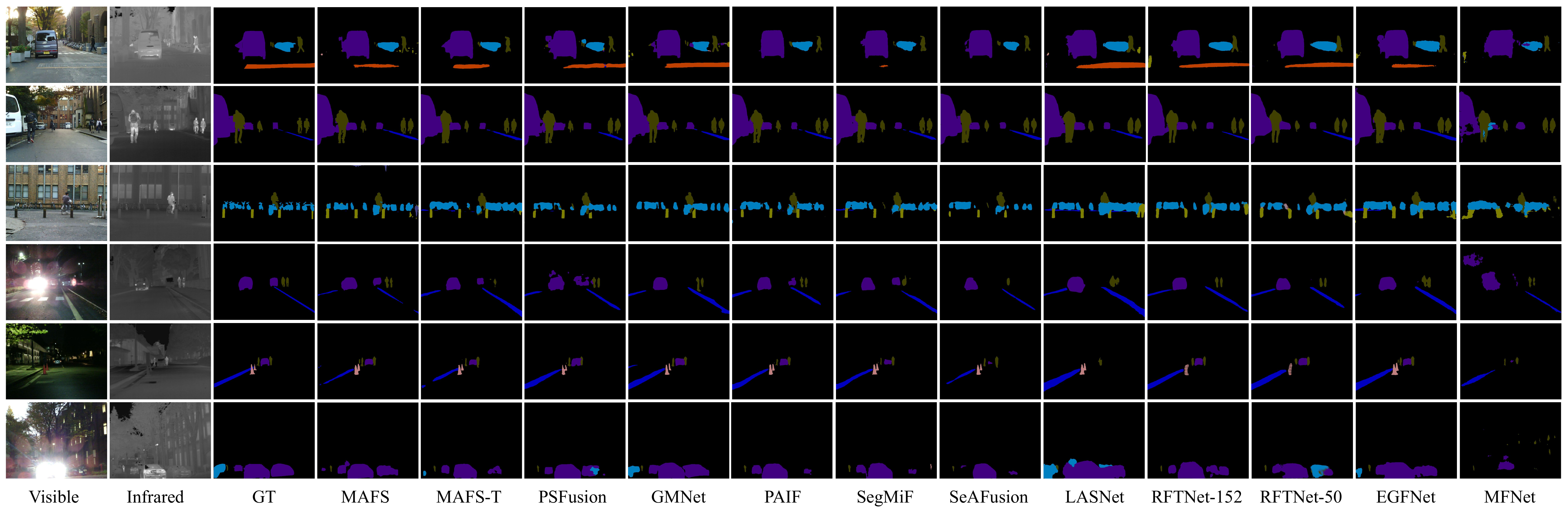}
    \caption{Visual comparisions in daytime and nighttime images of MFNet\cite{ha2017mfnet} dataset.}
    \label{MF_seg}
\end{figure*}

\begin{figure*}[h]
    \centering
    \includegraphics[width=0.7\linewidth]{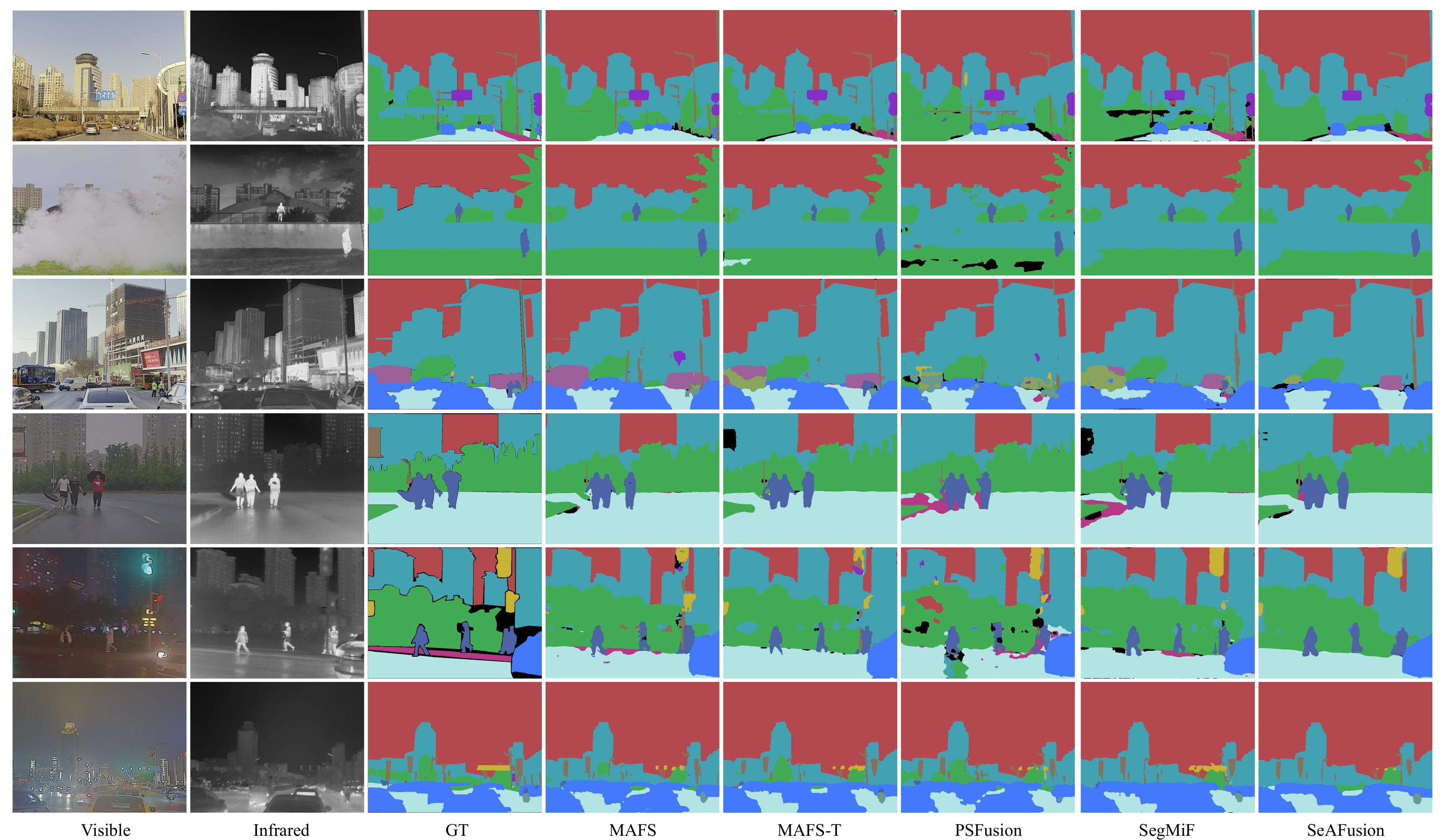}
    \caption{Visual comparisions in daytime and nighttime images of FMB\cite{liu2023multi} dataset.}
    \label{FMB_seg}
\end{figure*}

The segmentation color maps are shown intuitively in Fig. \ref{MF_seg} and Fig. \ref{FMB_seg}. In the daytime, the previous collaborative methods (i.e., SeAFusion, SegMiF and PAIF) show limited effectiveness in segmenting the ``Bump" category. Different fusion methods exhibit varying preferences for infrared and visible images. The FMB dataset encompasses diverse road scenes under different illumination conditions for scene parsing. In the daytime, MAFS enables the segmentation of significant categories, e.g., signs, cars on the road and people in heavy smog. The vegetation has more refined and clearer edges. Buses and trucks are correctly distinguished on the crowded roads. In the nighttime, MAFS enables segmenting the lamp category more accurately in the last two rows of Fig. \ref{FMB_seg}. MAFS and SegMiF better distinguish the boundaries between buildings and vegetation categories, which play a pivotal role in scene parsing with multiple categories. 

\begin{table*}[!ht]
    \centering
    \caption{Quantitative results (\%) of post-fusion segmentation strategy compared with fifteen SOTA fusion methods on the FMB\cite{liu2023multi} dataset.  (\textcolor{red}{\textbf{Red}}: best; \textcolor{blue}{\textbf{Blue}}: second best.)}
    \label{FMB_segtab2}
    \resizebox{0.7\textwidth}{!}{
    \begin{tabular}{c|c c c c c c c c|c}
    \toprule
        \textbf{Method} & \textbf{Car} & \textbf{Person} & \textbf{Truck} & \textbf{T-Lamp} & \textbf{T-Sign} & \textbf{Building} & \textbf{Vegetation} & \textbf{Pole} & \textbf{mIoU} \\ \midrule
        Visible & 78.07 & 55.75 & 45 & 35.03 & 75.13 & 81.14 & 85.46 & 49.87 & 59.08 \\ 
        Infrared & 74.35 & 69.15 & 29.3 & 27.46 & 61.97 & 81.68 & 81.93 & 33.88 & 55.08 \\ \midrule
        LRRNet\cite{li2023lrrnet} & 76.08 & 68.07 & 23.06 & 43.33 & 74.23 & 84.43 & 86.33 & 50.84 & 60.01 \\ 
        DeFusion\cite{liang2022fusion} & 80.92 & 70.7 & 26.3 & 44.51 & 75.46 & \textcolor{red}{\textbf{85.81}} & 86.92 & 52.01 & 60.6 \\ 
        TarDAL\cite{liu2022target} & 80.68 & 70.04 & 46.13 & 45.2 & 75.26 & 83.8 & \textcolor{blue}{\textbf{87.19}} & 50.38 & 62.25 \\ 
        DDFM\cite{zhao2023ddfm} & 78.94 & 70.57 & 28.92 & 42.74 & \textcolor{red}{\textbf{76.48}} & 84.3 & 85.48 & \textcolor{red}{\textbf{52.73}} & 61.93 \\ 
        PAIF\cite{liu2023paif} & 80.41 & 69.93 & \textcolor{red}{\textbf{53.63}} & 37.85 & 73.25 & 83.1 & \textcolor{red}{\textbf{87.88}} & \textcolor{blue}{\textbf{52.37}} & 61.31 \\ 
        SeAFusion\cite{tang2022image} & 79.82 & \textcolor{blue}{\textbf{70.71}} & 43.32 & 41.25 & 75 & 83.59 & 86.57 & 51.81 & 59.52 \\ 
        CDDFuse\cite{zhao2023cddfuse} & 80.43 & 68.08 & 39.12 & 35.28 & 74.48 & 82.79 & 86.46 & 46.31 & 61.55 \\ 
        EMMA\cite{zhao2024equivariant} & 79.17 & 69.38 & 46.37 & 36.79 & \textcolor{blue}{\textbf{75.98}} & 83.12 & 85.49 & 49.97 & 60.85 \\ 
        SHIP\cite{zheng2024probing} & 77.95 & \textcolor{red}{\textbf{70.72}} & 48.07 & 39.83 & 74.6 & 83.16 & 86.73 & 51.19 & 61.39 \\ 
        SegMiF\cite{liu2023multi} & \textcolor{red}{\textbf{81.06}} & 67.3 & 40 & 45.06 & 74.93 & 83.6 & 86.42 & 50.46 & 63.03 \\ 
        PSFusion\cite{tang2023rethinking} & 78.98 & 69.95 & 44.13 & \textcolor{blue}{\textbf{48.04}} & 74.04 & 84.37 & 87.11 & 52.28 & 63.14 \\ 
        BDLFusion\cite{liu2023bilevel} & 79.51 & 68.57 & 50.64 & 43.17 & 74.9 & \textcolor{blue}{\textbf{84.65}} & 85.93 & 50.11 & \textcolor{blue}{\textbf{63.75}} \\ 
        TIMFusion\cite{TIMFusion} & 80.07 & 67.48 & 35.79 & 40.7 & 74.55 & 82.02 & 86.15 & 48.94 & 61.6 \\
        CAF\cite{CAF} & 79.03 & 69.6 & 42.49 & 42.53 & 74.66 & 83.82 & 86.77 & 50.68 & 62.55 \\
        SAGE\cite{wu2025every} & \textcolor{blue}{\textbf{81.03}} & 68.83 & 35.52 & 43.93 & 71.6 & 83.86 & 85.35 & 47.64 & 62.75 \\ \midrule
        MAFS & 80.27 & 69.43 & \textcolor{blue}{\textbf{51.07}} & \textcolor{red}{\textbf{52.53}} & 74.46 & 84.51 & 86.05 & 49.64 & \textcolor{red}{\textbf{64.17}} \\ \bottomrule
    \end{tabular}
    }
\end{table*}

\begin{table*}[h!]
\centering
\caption{Quantitative results (\%) on the MFNet\cite{ha2017mfnet} dataset. The value 0.0 means that there are no true positives. MAFS-T indicates the teacher model. (\textcolor{red}{\textbf{Red}}: best; \textcolor{blue}{\textbf{Blue}}: second best.)}
\label{MF_segtab}
\resizebox{0.7\textwidth}{!}{
\begin{tabular}{c|cccccccc|c}
\toprule
\textbf{Method} & \textbf{Car} & \textbf{Person} & \textbf{Bike} & \textbf{Curve} & \textbf{Car Stop} & \textbf{Guardrail} & \textbf{Cone} & \textbf{Bump} & \textbf{mIoU} \\ \midrule
MFNet\cite{ha2017mfnet} & 65.9 & 58.9 & 42.9 & 29.9 & 9.9 & 0.0 & 25.2 & 27.7 & 39.7 \\
EGFNet\cite{dong2023egfnet} & 87.6 & 69.8 & 58.2 & 42.8 & 33.8 & 7.0 & 48.3 & 47.1 & 51.1 \\ 
RTFNet-50\cite{sun2019rtfnet} & 86.3 & 67.8 & 58.2 & \textcolor{blue}{\textbf{43.7}} & 24.3 & 3.6 & 26.0  & 57.2 & 51.7 \\
RTFNet-152\cite{sun2019rtfnet} & 87.4 & 70.3 & 62.7 & \textcolor{red}{\textbf{45.3}} & 29.8 & 0.0 & 29.1 & 55.7 & 53.2 \\
LASNet\cite{li2022rgb} & 84.2 & 67.1 & 56.9 & 41.1 & \textcolor{blue}{\textbf{39.6}} & \textcolor{red}{\textbf{18.9}} & 48.8 & 40.1 & 54.9 \\ 
SeAFusion\cite{tang2022image} & 84.1 & 70.6 & 57.1 & 35.8 & 19.1 & 0.0 & 44.4 & 32.8 & 49.1 \\ 
SegMiF\cite{liu2023multi} & 86.7  & 71.5  & \textcolor{blue}{\textbf{63.0}}  & 43.2  & 31.7  & 0.0  & \textcolor{red}{\textbf{56.7}}  & 52.4  & 55.9 \\ 
PAIF\cite{liu2023paif} & \textcolor{blue}{\textbf{88.1}}  & \textcolor{blue}{\textbf{72.4}}  & 60.8  & 39.8  & 29.5  & 6.5  & \textcolor{blue}{\textbf{56.0}}  & 57.2  & 56.5 \\
GMNet\cite{zhou2021gmnet} & 86.5 & \textcolor{red}{\textbf{73.1}} & 61.7 & 44 & \textcolor{red}{\textbf{42.3}} & 14.5 & 48.7 & 47.4 & 57.3 \\
PSFusion\cite{tang2023rethinking} & 82.6  & 71.0  & 60.6  & 40.6  & 27.6  & 7.1  & 37.0  & 53.2  & 53.0 \\
MAFS-T & \textcolor{red}{\textbf{89.3}}  & 72.0  & \textcolor{red}{\textbf{63.7}}  & 43.0  & 33.6  & 11.4  & 52.3  & \textcolor{blue}{\textbf{57.9}}  & \textcolor{red}{\textbf{57.9}} \\
MAFS & 86.3  & \textcolor{blue}{\textbf{72.4}}  & 59.5  & 41.6  & 38.1  & \textcolor{blue}{\textbf{15.3}}  & 45.3  & \textcolor{red}{\textbf{61.5}}  & \textcolor{blue}{\textbf{57.6}} \\
\bottomrule
\end{tabular}
}
\end{table*}

\begin{table*}[h!]
\centering
\caption{Quantitative results (\%) on the PST900\cite{shivakumar2020pst900} dataset. MAFS-T indicates the teacher model. (\textcolor{red}{\textbf{Red}}: best; \textcolor{blue}{\textbf{Blue}}: second best.)}
\label{PST_segtab}
\resizebox{0.7\textwidth}{!}{
\begin{tabular}{c|ccccc|c}
\toprule
\textbf{Method} & \textbf{Background} & \textbf{Hand-Drill} & \textbf{Backpack} & \textbf{Fire-Extinguisher}  & \textbf{Survivor} & \textbf{mIoU} \\ \midrule
ACNet\cite{hu2019acnet} & 99.25 & 51.46 & 83.19 & 59.95 & 65.19 & 71.81 \\
SA-Gate\cite{chen2020bi} & 99.25 & \textcolor{blue}{\textbf{81.01}} & 79.77 & 72.97 & 62.22 & 79.05 \\
MFNet\cite{ha2017mfnet} & 98.63 & 41.13 & 64.27 & 60.35 & 20.70 & 57.02 \\
PSTNet\cite{shivakumar2020pst900} & 98.85 & 53.60 & 69.20 & 70.12 & 50.03 & 68.36 \\
MFFENet\cite{zhou2021mffenet} & 99.40 & 72.50 & 81.02 & 66.38 & 75.60 & 78.98 \\
EGFNet\cite{dong2023egfnet} &  99.26 & 64.67 & 83.05 & 71.29 & 74.30 & 78.51 \\
MTANet\cite{zhou2022mtanet} & 99.33 & 62.05 & \textcolor{blue}{\textbf{87.50}} & 64.95 & \textcolor{red}{\textbf{79.14}} & 78.60 \\
FDCNet\cite{zhao2022feature} & 99.15 & 70.36 & 72.17 & 71.52 & 72.36 & 77.11 \\
MMSMCNet\cite{zhou2023mmsmcnet} & 99.39 & 62.36 & \textcolor{red}{\textbf{89.22}} & 73.29 & 74.70 & 79.80 \\
PSFusion\cite{tang2023rethinking} & 98.92 & 50.82 & 71.95 & 42.28 & 36.69 & 60.13 \\
MAFS-T & \textcolor{blue}{\textbf{99.44}}  & \textcolor{red}{\textbf{86.08}}  & 80.68  & \textcolor{red}{\textbf{78.72}}  & 76.55  & \textcolor{red}{\textbf{84.33}} \\
MAFS & \textcolor{red}{\textbf{99.49}} & 63.34 & 83.25 & \textcolor{blue}{\textbf{76.96}} & \textcolor{blue}{\textbf{77.61}} & \textcolor{blue}{\textbf{80.13}} \\ \bottomrule
\end{tabular}
}
\end{table*}

\subsubsection{Objective Comparision}
The quantitative assessment on MFNet, PST900, and FMB datasets is reported in Table \ref{MF_segtab}, \ref{PST_segtab}, and \ref{FMB_segtab}. 
MAFS surpasses GMNet 0.3\% in mIoU on the MFNet dataset. 
Among the eight categories, MAFS has remarkable performance in the guardrail and bump, demonstrating excellence in recognizing narrow elongated lines and small objects. For the PST900 dataset, our proposed method outperforms the second-best MMSMCNet by 0.33\% in mIoU. MAFS exhibits performance improvements in the ``Fire-Extinguisher'' and ``Survivor'' categories. MAFS is also compared with SeAFusion, PSFusion, and SegMiF on the FMB dataset. Shown in Table \ref{FMB_segtab}, our proposed collaborative training methods surpass the second-best SegMiF 1.6\% in mIoU. MAFS performs the best in the categories car and person, outperforming SegMiF 0.1\% mIoU in the car and 1.4\% mIoU in the person, respectively. The objective comparisons also illustrate MAFS enabling to distinguish the highlighted objects from cluttered backgrounds in complex scenarios, on account of context information transition and multilabel supervision. In Section \ref{ablation}, we further elaborate on the efficacy of the proposed architecture and training strategies.

\begin{table*}[!ht]
    \centering
    \caption{Quantitative results (\%) on the FMB\cite{liu2023multi} dataset. MAFS-T indicates the teacher model. (\textcolor{red}{\textbf{Red}}: best; \textcolor{blue}{\textbf{Blue}}: second best.)}
    \label{FMB_segtab}
    \resizebox{0.7\textwidth}{!}{
    \begin{tabular}{c|c c c c c c c c|c}
    \toprule
        \textbf{Method} & \textbf{Car} & \textbf{Person} & \textbf{Truck} & \textbf{T-Lamp} & \textbf{T-Sign} & \textbf{Building} & \textbf{Vegetation} & \textbf{Pole} & \textbf{mIoU} \\ \midrule
        SeAFusion\cite{tang2022image} & 76.7 & 64.6 & 8.5 & 27 & 70.1 & 80.7 & 84.2 & 42 & 53.9 \\ 
        SegMiF\cite{liu2023multi} & 78.3 & 65.4 & \textcolor{blue}{\textbf{47.3}} & \textcolor{red}{\textbf{43.1}} & \textcolor{red}{\textbf{74.8}} & \textcolor{blue}{\textbf{82}} & \textcolor{blue}{\textbf{85}} & \textcolor{blue}{\textbf{49.8}} & 54.8 \\ 
        PSFusion\cite{tang2023rethinking} & 75.3  & 57.3  & 20.4  & 22.5  & 70.0  & 80.9  & 84.7  & 38.6 & 50.3 \\
        MAFS-T & \textcolor{red}{\textbf{81.1}}  & \textcolor{red}{\textbf{70.1}}  & \textcolor{red}{\textbf{50.4}}  & \textcolor{blue}{\textbf{39.7}}  & \textcolor{blue}{\textbf{74.5}}  & \textcolor{red}{\textbf{84.9}}  & \textcolor{red}{\textbf{87.1}}  & \textcolor{red}{\textbf{52.4}}  & \textcolor{red}{\textbf{63.8}} \\
        MAFS & \textcolor{blue}{\textbf{78.4}} & \textcolor{blue}{\textbf{66.8}} & 24.3 & 26 & 71 & 81.9 & 84 & 46.4 & \textcolor{blue}{\textbf{56.4}} \\ \bottomrule
    \end{tabular}
    }
\end{table*}

\subsection{Ablation Study ($\mathbf{RQ3}$ and $\mathbf{RQ4}$)}
\label{ablation}
\subsubsection{effectiveness of blocks} For \textbf{model I} in Table \ref{architecture}, without the benefits from the semantic information and specific-task decoders, let fusion and segmentation networks share the same shallow feature extraction module and the overall network is optimized in the knowledge distillation fashion. 
In contrast, \textbf{model II} represents that the PHF is integrated into model I to evaluate the impact of the progressive fusion strategy. From observation between I and II, although PHF aggregates the shallow features and semantic features progressively, the performance of fusion and segmentation both suffer from a decline. To assess the decoders tailored to specific tasks, quantitative results are listed as III and IV in Table \ref{architecture}. \textbf{Model III} is the network getting rid of the transformer decoder for segmentation, and \textbf{model IV} represents the network without the fusion decoder. In comparison, two tasks require customized decoders to preserve the extracted rich features. The proposed fusion decoder protects rich features from information loss issues, in which INN fulfills the forward process with less information loss and the dense connection leads to more efficient learning and improved generalization. The segmentation decoder models long-range global dependencies, which enlarge the receptive fields and keep high-resolution deep semantic features to aggregate multi-scale feature maps and deliver more accurate segmentation results. To verify the viewpoint of heterogeneous features, i.e., shallow features and semantic features oriented for two tasks in Section \ref{shallow_deep_features}, qualitative results are listed from V to IX in Table \ref{architecture}. \textbf{Model V} indicates that PHF achieves the progressive fusion from $\mathcal{F}^{i}$ deep semantic features at index $i$ for $i=3, 4, 5, 6$ to $\mathcal{F}^{i}$ shallow features at index $i$ for  $i=1, 2$. \textbf{Model VI} represents the PHF utilizing $\mathcal{F}^{i}$ deep semantic features at index $i$ for $i=3, 4, 5$ and shallow features. \textbf{Model VII} denotes utilizing $\mathcal{F}^{i}$ deep semantic features at index $i$ for $i=3, 4$ and the shallow features to achieve PHF. In \textbf{model VIII}, only shallow features implement the fusion task. And \textbf{model IX} substitutes the PHF strategy with semantic deep features $\mathcal{F}^{3}$ as the intermediate features for image fusion to further investigate the impact of deep semantic features on image fusion performance. By comparing various fusion circumstances from V to VII reported in Table \ref{architecture}, the integration of deep semantic features enhances the visual quality of the fused images, especially the model VI, when task-specific decoders exhibit superior reconstruction performance. Concretely, model VIII suffers a noticeable degradation in performance on both tasks, as degradation in image fusion compromises the teacher model's ability to guide the training process in semantic segmentation. The quantitative results of model IX demonstrate the case that deep semantic features $\mathcal{F}^{3}$ are incompetent for the generation of fused images. It's the best choice to combine deep semantic features $\mathcal{F}^{3}$ and shallow features via the proposed PHF strategy.



\begin{table}[!ht]
    \centering
    \caption{Quantitative results about architecture design.}
    \label{architecture}
    \resizebox{\linewidth}{!}{
    \begin{tabular}{c|c c c c c c|c}
    \toprule
        \textbf{Models} & EN &  SD &  SF &  VIF &  Qabf &  AG & mIoU \\ \midrule
        I & 6.8248 & 41.6328 & 14.0795 & 1.0857 & 0.7158 & 6.5696 & 55.1 \\ 
        II & 6.8206 & 41.3449 & 14.5121 & 1.0713 & 0.7041 & 6.8171 & 54.4 \\ 
        III & \textbf{7.0037} & 46.1563 & 15.0914 & 1.1605 & 0.7091 & 7.081 & 55.0 \\
        IV &6.8036 & 40.5156 & 13.8834 & 1.0881 & 0.7192 & 6.5295 & 56.1 \\
        V &  6.9607 & 45.3601 & 14.315 & 1.1588 & 0.7201 & 6.7272 & 55.0 \\
        VI & 6.9584 & 44.434 & 14.2053 & \textbf{1.1654} & \textbf{0.7259} & 6.6638 & 56.0 \\
        VII & 6.9767 & 45.7541 & 14.6985 & 1.1587 & 0.7127 & 6.9077 & 55.9\\
        VIII & 6.8628 & 42.932 & 14.1267 & 1.1046 & 0.7157 & 6.6178 & 54.6\\
        IX & 6.4836 & 32.4578 & 9.2473 & 0.5377 & 0.389 & 4.6344 & 53.3\\
        Ours & 7.0033 & \textbf{46.6232} & \textbf{15.3121} & 1.1437 & 0.7102 & \textbf{7.1457} & \textbf{57.6}\\
        \bottomrule
    \end{tabular}
    }
\end{table}

\begin{figure}[h]
    \centering
    \includegraphics[width=0.9\linewidth]{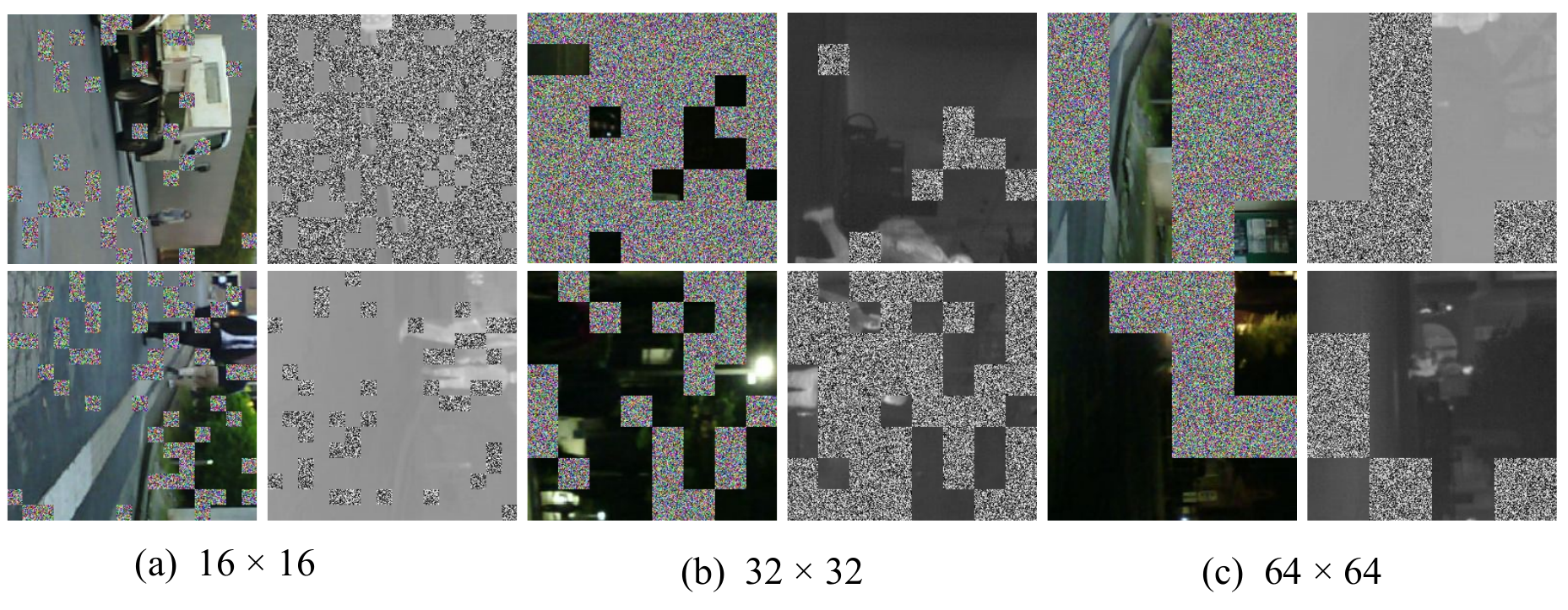}
    \caption{Visual comparisons of paired 256 $\times$ 256 visible and infrared images with different masking sizes. And the size of 32 $\times$ 32 is the optimal.}
    \label{mask_size_pic}
\end{figure}

\begin{table}[!ht]
    \centering
    \caption{Quantitative results about training with different masking strategies and dynamic balancing parameters.}
    \label{mask_size}
    \resizebox{\linewidth}{!}{
    \begin{tabular}{c|c c c c c c c}
    \toprule
        \textbf{Mask size} & EN &  SD &  SF &  VIF &  Qabf &  \multicolumn{1}{c|}{ AG} & mIoU \\ \midrule
        $16\times 16$ & 6.9438 & 43.3705 & 14.4310 & 1.1270 & \textbf{0.7105} & \multicolumn{1}{c|}{ 6.8441} & 56.0 \\ 
        $64\times64$ & \textbf{7.0100} & 45.4855 & 15.3019 & \textbf{1.1539} & 0.6996 & \multicolumn{1}{c|}{\textbf{7.2460}} & 55.0 \\
        $32\times32$ (Ours) & 7.0033 & \textbf{46.6232} & \textbf{15.3121} & 1.1437 & 0.7102 & \multicolumn{1}{c|}{7.1457} & \textbf{57.6}\\\midrule
         \textbf{Mask ratio} & ~ &  ~ &  ~ & ~ & ~ &  ~ & ~ \\ \midrule
        0.5 & 6.9818 & 46.1826 & 14.863 & 1.1426 & 0.7101 & \multicolumn{1}{c|}{6.9740} & 55.3 \\ 
        0.7 & \textbf{7.0390} & 46.4374 & 15.2021 & \textbf{1.1558} & 0.7093 & \multicolumn{1}{c|}{7.1261} & 54.7 \\
         Random (Ours) & 7.0033 & \textbf{46.6232} & \textbf{15.3121} & 1.1437 & \textbf{0.7102} & \multicolumn{1}{c|}{\textbf{7.1457}} & \textbf{57.6}\\\midrule
        \textbf{Strategy} & ~ & ~ &  ~ &  ~ & ~ & ~ &  ~ \\ \midrule
        EW & \textbf{7.0774} & \textbf{46.9425} & \textbf{15.3204} & 1.1161 & 0.6887 & \multicolumn{1}{c|}{\textbf{7.2922}} & 54.0 \\
        UW\cite{kendall2018multi} & 7.0125 & 46.73 & 15.2679 & \textbf{1.1486} & \textbf{0.711} & \multicolumn{1}{c|}{7.1861} & 55.3\\
        FairGrad\cite{ban2024fair} & 7.0519 & 45.9164 & 14.869 & 1.1197 & 0.6953 & \multicolumn{1}{c|}{7.0264} & 52.1 \\
        $w/o~\alpha$-fair\cite{liu2019end} & 6.8957 & 44.1628 & 14.3795 & 1.0960 & 0.7076 & \multicolumn{1}{c|}{6.7514} & 55.1 \\ 
        $\alpha~$=~0.5 & 6.9870 & 45.2053 & 14.8603 & 1.1273 & 0.7083 & \multicolumn{1}{c|}{7.0216} & 57.0 \\ 
        $\alpha~$=~0.8 (Ours) & 7.0033 & 46.6232 & 15.3121 & 1.1437 & 0.7102 & \multicolumn{1}{c|}{7.1457} & \textbf{57.6}\\
         \bottomrule
    \end{tabular}
    }
\end{table}

\subsubsection{effectiveness of training strategy}As for the masking strategy, we discuss the sizes of masking patches and masking ratios. Quantitative results are listed in Table \ref{mask_size}. Masking strategies (as shown in Fig. \ref{mask_size_pic}) in the reconstruction stage play a positive role in extracting abundant representations and further promoting the final image fusion performance. To investigate influences on masking regions, we change the \textbf{Mask size} to 16$\times$16 and 64$\times$64. Intuitively, they both get better fusion outcomes, but suffer a sharp decline in segmentation performance. Relatively, results about \textbf{Mask ratio} in Table \ref{mask_size} demonstrate that a fixed masking ratio restricts the model from capturing diverse information. Therefore, we utilize the mask size of 32$\times$32 and a random masking ratio to train an encoder with stronger feature extraction capability, which is better suited for multi-task scenarios. 
\label{sec4_DWA}

\begin{figure}[h]
    \centering
    \includegraphics[width=0.9\linewidth]{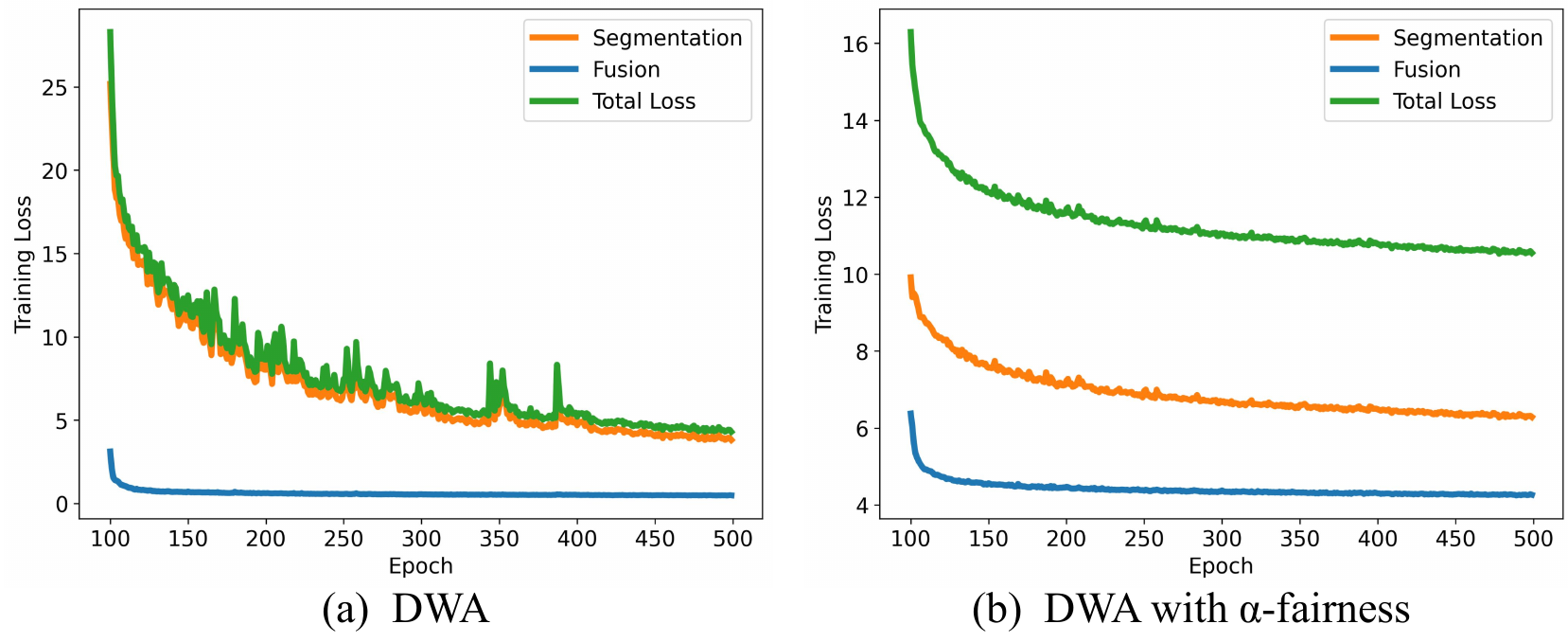}
    \caption{Visual comparisons of DWA and $\alpha$-fairness during the training stage II. $\alpha$ is set to 0.8.}
    \label{dynamic_weight}
\end{figure}

To better optimize the proposed model, we introduce $\alpha$ fairness into the dynamic weight factor as the learnable parameters for Eq. (\ref{15}). Fairness is closely related to resource allocation in wireless communication \cite{ban2024fair}. Max-min fairness emphasizes providing the user who receives the least resources with the maximum possible allocation. In our unified framework, semantic segmentation is the harder-to-train task, hence max-min fairness with a larger $\alpha$ is expected. We provide results about \textbf{Strategy:} utilizing different balancing ways, i.e., equal weighting (EW), uncertainty weighting (UW) \cite{kendall2018multi}, FairGrad\cite{ban2024fair}, and dynamic weight average (DWA)\cite{liu2019end} in Table \ref{mask_size}. Concretely, the fused images generated by assigning EW to the task losses exhibit high contrast. UW yields the best visual fidelity. FairGrad struggles to resolve the gradient conflicts between the two tasks. Compared with DWA \cite{liu2019end} without $\alpha$-fairness, when $\alpha$ is set to 0.5, the segmentation task yields convincing results, but the quantitative visual metrics of fusion show performance degradation.  As illustrated in Fig. \ref{dynamic_weight}, the losses trained with $\alpha$-fairness exhibit a smoother and more stable decrease compared to DWA. Therefore, we configure $\alpha$ as 0.8 empirically and strike a balance between the two tasks with different training difficulties.



\begin{table}[!ht]
    \centering
    \caption{Quantitative results about loss functions design in stage II.}
    \label{loss_function}
    \resizebox{\linewidth}{!}{
    \begin{tabular}{>{\centering\arraybackslash}p{0.4cm} >{\centering\arraybackslash}p{0.4cm} >{\centering\arraybackslash}p{0.4cm} >{\centering\arraybackslash}p{0.4cm}|c c c c c c|c}
    \toprule
         $\mathcal{L}_{fus}$ & $\mathcal{L}_{seg}$ & $\mathcal{L}_{aux}$ & $\mathcal{L}_{kd}$ &  EN &  SD &  SF &  VIF &  Qabf &  AG & mIoU \\ \midrule
        \checkmark & \checkmark & ~ & ~ & 6.9474 & 45.2449 & 14.5759 & 1.1359 & 0.7162 & 6.8416 & 53.2 \\
        \checkmark & \checkmark & \checkmark & ~ & 6.9179 & 43.7489 & 14.3141 & 1.1260 & 0.7206 & 6.7331 & 52.0 \\
        \checkmark & \checkmark & ~ & \checkmark & 6.8879 & 43.2003 & 14.0460 & 1.1288 & \textbf{0.7225} & 6.6207 & 55.6 \\
            \checkmark & \checkmark & \checkmark & \checkmark & \textbf{7.0033} & \textbf{46.6232} & \textbf{15.3121} & \textbf{1.1437} & 0.7102 & \textbf{7.1457} & \textbf{57.6} \\
        \bottomrule
    \end{tabular}
    }
\end{table}

\begin{table}[!ht]
    \centering
    \caption{Quantitative results about different segmentation backbone models. Left: Teacher mIoU, Right: Student mIoU.}
    \label{backbone_ablation}
    \resizebox{\linewidth}{!}{
    \begin{tabular}{c|c|c| c c c c c c|c}
    \toprule
        \textbf{Models} & \textbf{Backbone} & mIoU & EN &  SD &  SF &  VIF &  Qabf &  AG & mIoU \\ \midrule
        FeedFormer\cite{shim2023feedformer} & MiT-B2 & 55.1 &  7.0087 & 46.793 & 15.3944 & 1.1445 & 0.7064 & 7.2295 & 57.2\\
        SegFormer\cite{xie2021segformer} & MiT-B5 & 57.0 & \textbf{7.0664} & \textbf{47.941} & \textbf{15.511} & \textbf{1.1501} & 0.6936 & \textbf{7.3455} & 55.3 \\
        SegNeXt\cite{guo2022segnext} & MSCAN-L & \textbf{57.9}  & 7.0033 & 46.6232 & 15.3121 & 1.1437 & \textbf{0.7102} & 7.1457 & \textbf{57.6}\\
        \bottomrule
    \end{tabular}
    }
\end{table}

To retain texture details and high contrast, we introduce $\mathcal{L}_{int}$ and $\mathcal{L}_{grad}$ as the guidance for the image fusion task in Section \ref{sec_loss}. $\mathcal{L}_{seg}$ is applied for the segmentation task. $\mathcal{L}_{aux}$ is the auxiliary loss for the shape and contour supervision. $\mathcal{L}_{kd}$ is employed to align logits maps between teacher and student models. According to Table \ref{loss_function}, we observe that $\mathcal{L}_{kd}$ can excavate information related to segmentation from the teacher model. Nevertheless, $\mathcal{L}_{aux}$ interferes with segmentation performance, as the auxiliary loss increases the difficulty of optimizing the task-specific training loss. When combining them into the customized loss function, the unified framework enables the optimal balance between fusion and segmentation. This also demonstrates the rationality of the constraints we proposed.

To verify the robustness of the proposed framework, FeedFormer\cite{shim2023feedformer} and SegFormer\cite{xie2021segformer} are also pre-trained and employed as segmentation networks. The quantitative results are reported in Table \ref{backbone_ablation}. Image fusion is a relatively easy-to-train task compared to segmentation, especially with the help of semantic information injected from deep cross-modality features. It's vital to consider varying task difficulties and strike a balance between task-oriented loss functions. In terms of semantic segmentation, logits maps are treated as soft targets that carry more helpful information than one-hot labels, enabling the student segmentation network to effectively transfer segmentation capabilities from the teacher model and eliminate dependency during the inference phase.

\begin{figure}[h]
    \centering
    \includegraphics[width=0.95\linewidth]{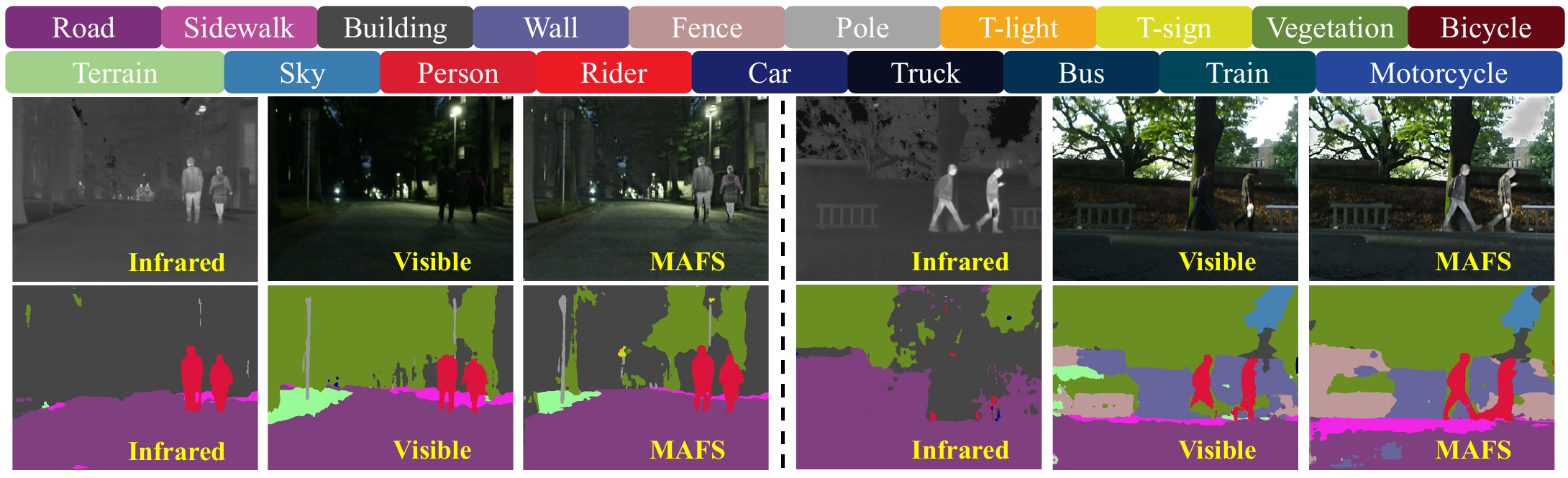}
    \caption{Generalization validation on the MFNet\cite{ha2017mfnet} dataset by pre-trained DeepLabV3+\cite{chen2018encoder}.}
    \label{generalization}
\end{figure}

\begin{table}[!ht]
    \centering
    \caption{Complexity evaluation of the distillation-based network MAFS. MAFS-S indicates the student network tailored for segmentation.}
    \label{distillation}
    \resizebox{0.8\linewidth}{!}{
    \begin{tabular}{c|c | c c| c c}
    \toprule
        \textbf{Models} & Input Size & Fusion & Segmentation & Flops (G) & Params (M) \\ \midrule
        SegNeXt\cite{guo2022segnext} & 256 $\times$ 256 & ~ & \checkmark & 16.17 & 48.78 \\ 
        MAFS-S & 640 $\times$ 480 & ~ & \checkmark & 221.08 & 128.41 \\ 
        MAFS & 640 $\times$ 480 & \checkmark & \checkmark & 324.22 & 128.69 \\ \bottomrule
    \end{tabular}
    }
\end{table}

\begin{table}[!ht]
    \centering
    \caption{Complexity evaluation of segmentation methods and collaborative-based methods on the MFNet test set.}
    \label{complexity}
    \resizebox{0.8\linewidth}{!}{
    \begin{tabular}{c| c c | c c}
    \toprule
        Model & Fusion & Segmentation & Flops(G) & Params(M) \\ \midrule
        LASNet\cite{li2022rgb} & ~ & \checkmark & 234.45 & 165.03 \\ 
        EGFNet\cite{dong2023egfnet} & ~ & \checkmark & 201.5 & 62.78 \\ 
        SeAFusion\cite{tang2022image} & \checkmark & \checkmark & 65.54 & 13.06 \\ 
        SegMiF\cite{liu2023multi} & \checkmark & \checkmark & 359.41 & 45.57 \\ 
        PSFusion\cite{tang2023rethinking} & \checkmark & \checkmark & 180.86 & 45.9 \\ 
        MAFS & \checkmark & \checkmark & 324.22 & 128.69 \\ \bottomrule
    \end{tabular}
    }
\end{table}
\subsection{Discussion}
\subsubsection{Generalizability and Complexity Analysis}
To validate the generalizability of MAFS, DeepLabV3+\cite{chen2018encoder} pre-trained on the Cityscapes dataset is introduced to segment objects in autonomous driving scenes. In Fig. \ref{generalization}, MAFS combined the imaging strengths of distinct modalities. The significant objects are segmented close to the segmentation results of visible images, which demonstrates the generalization ability of our model in real-world scenarios.
Owing to the guidance of pre-trained SegNeXt, 48.78M parameters are introduced during the training stage II as reported in Table \ref{distillation}. The additional computational cost is 16.17 GFlops. MAFS increases 103.14 GFlops and 0.28M parameters compared with the dedicated segmentation network MAFS-S during the inference phase. The complexity evaluation of segmentation and collaborative-based methods is listed in Table \ref{complexity}. By quantitative comparisons, SeAFusion achieved real-time fusion and segmentation with the least computational cost and network parameters. MAFS has a lower computational cost than that of SegMiF and fewer parameters than those of LASNet.
\subsubsection{Limitations and Outlook}
MAFS implements the collaborative training of infrared-visible image fusion and semantic segmentation under a unified network. There are still some issues that merit further attention. The lightweight network should be considered to better support real-world applications and deployment on mobile devices. In particular, a more effective backbone is expected to improve performance and reduce the complexity. Combined with the generative ability of the diffusion model, the multi-task framework has the potential for further development. It's also a prevailing trend to consider the out-of-distribution generalization problem associated with open-set learning. Last but not least, measuring and reducing the semantic gap between low-level and high-level tasks remains a critical challenge that calls for further research and improvement in future work.
\section{Concluding Remarks}
\label{sec5}
In this paper, we propose a two-stage training strategy for the joint learning of infrared-visible image fusion and semantic segmentation. MAE is trained to reconstruct source images as a pretext task. The teacher-student paradigm in the second stage fulfills collaborative training of two tasks. In concrete terms, the adaptive fusion module (AFM) fuses shallow features, and then the progressive heterogeneous fusion module (PHF) aggregates them with semantic features in a progressive way. Two task-specific decoders are employed to get fused images with less information loss and convincing segmentation results, respectively. The fairness-based dynamic factor is weighted on fusion and segmentation tasks for stable training. Experiments on fusion and segmentation datasets evaluate the efficacy of MAFS compared with state-of-the-art methods. Future work includes combining the diffusion model with the AE-based methods for infrared-visible image fusion.


\begin{thebibliography}{1}
\bibliographystyle{IEEEtran}

\bibitem{PIG}
Z.~Xie, R.~Qiu, S.~Wang, X.~Tan, Y.~Xie, and L.~Ma, ``Pig: Prompt images guidance for night-time scene parsing,'' \emph{IEEE Trans. Image Process.}, vol.~33, pp. 3921--3934, 2024.

\bibitem{zhao2017pyramid}
H.~Zhao, J.~Shi, X.~Qi, X.~Wang, and J.~Jia, ``Pyramid scene parsing network,'' in \emph{Proc. IEEE Conf. Comput. Vis. Pattern Recog.}, 2017, pp. 2881--2890.

\bibitem{ma2023overview}
S.~Ma, J.~Gao, R.~Wang, J.~Chang, Q.~Mao, Z.~Huang, and C.~Jia, ``Overview of intelligent video coding: from model-based to learning-based approaches,'' \emph{Vis. Intell.}, vol.~1, no.~1, p.~15, 2023.

\bibitem{liu2024coconet}
J.~Liu, R.~Lin, G.~Wu, R.~Liu, Z.~Luo, and X.~Fan, ``Coconet: Coupled contrastive learning network with multi-level feature ensemble for multi-modality image fusion,'' \emph{Int. J. Comput. Vis.}, vol. 132, no.~5, pp. 1748--1775, 2024.

\bibitem{li2023lrrnet}
H.~Li, T.~Xu, X.-J. Wu, J.~Lu, and J.~Kittler, ``Lrrnet: A novel representation learning guided fusion network for infrared and visible images,'' \emph{IEEE Trans. Pattern Anal. Mach. Intell.}, vol.~45, no.~9, pp. 11\,040--11\,052, 2023.

\bibitem{M2FNet}
X.~Li, S.~Chen, C.~Tian, H.~Zhou, and Z.~Zhang, ``M2fnet: Mask-guided multi-level fusion for rgb-t pedestrian detection,'' \emph{IEEE Trans. Multimedia}, vol.~26, pp. 8678--8690, 2024.

\bibitem{peng2023adaptive}
J.~Peng, G.~Jiang, and H.~Wang, ``Adaptive memorization with group labels for unsupervised person re-identification,'' \emph{IEEE Trans. Circuit Syst. Video Technol.}, vol.~33, no.~10, pp. 5802--5813, 2023.

\bibitem{li2018cross}
C.~Li, C.~Zhu, Y.~Huang, J.~Tang, and L.~Wang, ``Cross-modal ranking with soft consistency and noisy labels for robust rgb-t tracking,'' in \emph{Proc. Eur. Conf. Comput. Vis.}, 2018, pp. 808--823.

\bibitem{tang2023rethinking}
L.~Tang, H.~Zhang, H.~Xu, and J.~Ma, ``Rethinking the necessity of image fusion in high-level vision tasks: A practical infrared and visible image fusion network based on progressive semantic injection and scene fidelity,'' \emph{Inf. Fusion}, vol.~99, p. 101870, 2023.

\bibitem{liu2016image}
Y.~Liu, X.~Chen, R.~K. Ward, and Z.~J. Wang, ``Image fusion with convolutional sparse representation,'' \emph{IEEE Signal Processing Lett.}, vol.~23, no.~12, pp. 1882--1886, 2016.

\bibitem{zhao2020bayesian}
Z.~Zhao, S.~Xu, C.~Zhang, J.~Liu, and J.~Zhang, ``Bayesian fusion for infrared and visible images,'' \emph{Signal Process.}, vol. 177, p. 107734, 2020.

\bibitem{liang2022fusion}
P.~Liang, J.~Jiang, X.~Liu, and J.~Ma, ``Fusion from decomposition: A self-supervised decomposition approach for image fusion,'' in \emph{Proc. Eur. Conf. Comput. Vis.}, 2022, pp. 719--735.

\bibitem{xu2020u2fusion}
H.~Xu, J.~Ma, J.~Jiang, X.~Guo, and H.~Ling, ``U2fusion: A unified unsupervised image fusion network,'' \emph{IEEE Trans. Pattern Anal. Mach. Intell.}, vol.~44, no.~1, pp. 502--518, 2020.

\bibitem{ma2022swinfusion}
J.~Ma, L.~Tang, F.~Fan, J.~Huang, X.~Mei, and Y.~Ma, ``Swinfusion: Cross-domain long-range learning for general image fusion via swin transformer,'' \emph{IEEE/CAA J. Autom. Sinica}, vol.~9, no.~7, pp. 1200--1217, 2022.

\bibitem{AFT}
Z.~Chang, Z.~Feng, S.~Yang, and Q.~Gao, ``Aft: Adaptive fusion transformer for visible and infrared images,'' \emph{IEEE Trans. Image Process.}, vol.~32, pp. 2077--2092, 2023.

\bibitem{Dif-Fusion}
J.~Yue, L.~Fang, S.~Xia, Y.~Deng, and J.~Ma, ``Dif-fusion: Toward high color fidelity in infrared and visible image fusion with diffusion models,'' \emph{IEEE Trans. Image Process.}, vol.~32, pp. 5705--5720, 2023.

\bibitem{zhao2023ddfm}
Z.~Zhao, H.~Bai, Y.~Zhu, J.~Zhang, S.~Xu, Y.~Zhang, K.~Zhang, D.~Meng, R.~Timofte, and L.~Van~Gool, ``Ddfm: denoising diffusion model for multi-modality image fusion,'' in \emph{Proc. IEEE Int. Conf. Comput. Vis.}, 2023, pp. 8082--8093.

\bibitem{Different}
H.~Li, Y.~Cen, Y.~Liu, X.~Chen, and Z.~Yu, ``Different input resolutions and arbitrary output resolution: A meta learning-based deep framework for infrared and visible image fusion,'' \emph{IEEE Trans. Image Process.}, vol.~30, pp. 4070--4083, 2021.

\bibitem{tang2022image}
L.~Tang, J.~Yuan, and J.~Ma, ``Image fusion in the loop of high-level vision tasks: A semantic-aware real-time infrared and visible image fusion network,'' \emph{Inf. Fusion}, vol.~82, pp. 28--42, 2022.

\bibitem{sun2022detfusion}
Y.~Sun, B.~Cao, P.~Zhu, and Q.~Hu, ``Detfusion: A detection-driven infrared and visible image fusion network,'' in \emph{ACM Int. Conf. Multimedia}, 2022, pp. 4003--4011.

\bibitem{liu2022target}
J.~Liu, X.~Fan, Z.~Huang, G.~Wu, R.~Liu, W.~Zhong, and Z.~Luo, ``Target-aware dual adversarial learning and a multi-scenario multi-modality benchmark to fuse infrared and visible for object detection,'' in \emph{Proc. IEEE Conf. Comput. Vis. Pattern Recognit.}, 2022, pp. 5802--5811.

\bibitem{liu2023multi}
J.~Liu, Z.~Liu, G.~Wu, L.~Ma, R.~Liu, W.~Zhong, Z.~Luo, and X.~Fan, ``Multi-interactive feature learning and a full-time multi-modality benchmark for image fusion and segmentation,'' in \emph{Proc. IEEE Int. Conf. Comput. Vis.}, 2023, pp. 8115--8124.

\bibitem{liu2023paif}
Z.~Liu, J.~Liu, B.~Zhang, L.~Ma, X.~Fan, and R.~Liu, ``Paif: Perception-aware infrared-visible image fusion for attack-tolerant semantic segmentation,'' in \emph{ACM Int. Conf. Multimedia}, 2023, pp. 3706--3714.

\bibitem{liu2023bilevel}
Z.~Liu, J.~Liu, G.~Wu, L.~Ma, X.~Fan, and R.~Liu, ``Bi-level dynamic learning for jointly multi-modality image fusion and beyond,'' \emph{IJCAI}, 2023.

\bibitem{zhang2020rethinking}
H.~Zhang, H.~Xu, Y.~Xiao, X.~Guo, and J.~Ma, ``Rethinking the image fusion: A fast unified image fusion network based on proportional maintenance of gradient and intensity,'' in \emph{Proc. AAAI Conf. Artif. Intell.}, vol.~34, no.~07, 2020, pp. 12\,797--12\,804.

\bibitem{zhang2021sdnet}
H.~Zhang and J.~Ma, ``Sdnet: A versatile squeeze-and-decomposition network for real-time image fusion,'' \emph{Int. J. Comput. Vis.}, vol. 129, no.~10, pp. 2761--2785, 2021.

\bibitem{CHITNet}
K.~Du, H.~Li, Y.~Zhang, and Z.~Yu, ``Chitnet: A complementary to harmonious information transfer network for infrared and visible image fusion,'' \emph{IEEE Trans. Instrum. Meas.}, vol.~74, pp. 1--17, 2025.

\bibitem{ma2019fusiongan}
J.~Ma, W.~Yu, P.~Liang, C.~Li, and J.~Jiang, ``Fusiongan: A generative adversarial network for infrared and visible image fusion,'' \emph{Inf. Fusion}, vol.~48, pp. 11--26, 2019.

\bibitem{liu2021learning}
J.~Liu, X.~Fan, J.~Jiang, R.~Liu, and Z.~Luo, ``Learning a deep multi-scale feature ensemble and an edge-attention guidance for image fusion,'' \emph{IEEE Trans. Circuit Syst. Video Technol.}, vol.~32, no.~1, pp. 105--119, 2021.

\bibitem{yi2024diff}
X.~Yi, L.~Tang, H.~Zhang, H.~Xu, and J.~Ma, ``Diff-if: Multi-modality image fusion via diffusion model with fusion knowledge prior,'' \emph{Inf. Fusion}, vol. 110, p. 102450, 2024.

\bibitem{li2018densefuse}
H.~Li and X.-J. Wu, ``Densefuse: A fusion approach to infrared and visible images,'' \emph{IEEE Trans. Image Process.}, vol.~28, no.~5, pp. 2614--2623, 2018.

\bibitem{zhao2023cddfuse}
Z.~Zhao, H.~Bai, J.~Zhang, Y.~Zhang, S.~Xu, Z.~Lin, R.~Timofte, and L.~Van~Gool, ``Cddfuse: Correlation-driven dual-branch feature decomposition for multi-modality image fusion,'' in \emph{Proc. IEEE Conf. Comput. Vis. Pattern Recognit.}, 2023, pp. 5906--5916.

\bibitem{zhao2024equivariant}
Z.~Zhao, H.~Bai, J.~Zhang, Y.~Zhang, K.~Zhang, S.~Xu, D.~Chen, R.~Timofte, and L.~Van~Gool, ``Equivariant multi-modality image fusion,'' in \emph{Proc. IEEE Conf. Comput. Vis. Pattern Recognit.}, 2024, pp. 25\,912--25\,921.

\bibitem{liu2025}
J.~Liu, G.~Wu, Z.~Liu, D.~Wang, Z.~Jiang, L.~Ma, W.~Zhong, X.~Fan, and R.~Liu, ``Infrared and visible image fusion: From data compatibility to task adaption,'' \emph{IEEE Trans. Pattern Anal. Mach. Intell.}, vol.~47, no.~4, pp. 2349--2369, 2025.

\bibitem{li2024deep}
H.~Li, J.~Liu, Y.~Zhang, and Y.~Liu, ``A deep learning framework for infrared and visible image fusion without strict registration,'' \emph{Int. J. Comput. Vis.}, vol. 132, no.~5, pp. 1625--1644, 2024.

\bibitem{MulFS-CAP}
H.~Li, Z.~Yang, Y.~Zhang, W.~Jia, Z.~Yu, and Y.~Liu, ``Mulfs-cap: Multimodal fusion-supervised cross-modality alignment perception for unregistered infrared-visible image fusion,'' \emph{IEEE Trans. Pattern Anal. Mach. Intell.}, vol.~47, no.~5, pp. 3673--3690, 2025.

\bibitem{tang2022superfusion}
L.~Tang, Y.~Deng, Y.~Ma, J.~Huang, and J.~Ma, ``Superfusion: A versatile image registration and fusion network with semantic awareness,'' \emph{IEEE/CAA J. Autom. Sinica}, vol.~9, no.~12, pp. 2121--2137, 2022.

\bibitem{wang2023interactively}
D.~Wang, J.~Liu, R.~Liu, and X.~Fan, ``An interactively reinforced paradigm for joint infrared-visible image fusion and saliency object detection,'' \emph{Inf. Fusion}, vol.~98, p. 101828, 2023.

\bibitem{yang2025instruction}
Z.~Yang, Y.~Zhang, H.~Li, and Y.~Liu, ``Instruction-driven fusion of infrared--visible images: Tailoring for diverse downstream tasks,'' \emph{Inf. Fusion}, vol. 121, p. 103148, 2025.

\bibitem{CAF}
J.~Liu, G.~Wu, Z.~Liu, L.~Ma, R.~Liu, and X.~Fan, ``Where elegance meets precision: Towards a compact, automatic, and flexible framework for multi-modality image fusion and applications,'' \emph{IJCAI}, 2024.

\bibitem{TIMFusion}
R.~Liu, Z.~Liu, J.~Liu, X.~Fan, and Z.~Luo, ``A task-guided, implicitly-searched and meta-initialized deep model for image fusion,'' \emph{IEEE Trans. Pattern Anal. Mach. Intell.}, vol.~46, no.~10, pp. 6594--6609, 2024.

\bibitem{wu2025every}
G.~Wu, H.~Liu, H.~Fu, Y.~Peng, J.~Liu, X.~Fan, and R.~Liu, ``Every sam drop counts: Embracing semantic priors for multi-modality image fusion and beyond,'' in \emph{Proc. IEEE Conf. Comput. Vis. Pattern Recognit.}, 2025, pp. 17\,882--17\,891.

\bibitem{kirillov2023segment}
A.~Kirillov, E.~Mintun, N.~Ravi, H.~Mao, C.~Rolland, L.~Gustafson, T.~Xiao, S.~Whitehead, A.~C. Berg, W.-Y. Lo \emph{et~al.}, ``Segment anything,'' in \emph{Proc. IEEE Int. Conf. Comput. Vis.}, 2023, pp. 4015--4026.

\bibitem{chen2018encoder}
L.-C. Chen, Y.~Zhu, G.~Papandreou, F.~Schroff, and H.~Adam, ``Encoder-decoder with atrous separable convolution for semantic image segmentation,'' in \emph{Proc. Eur. Conf. Comput. Vis.}, 2018, pp. 801--818.

\bibitem{wang2020deep}
J.~Wang, K.~Sun, T.~Cheng, B.~Jiang, C.~Deng, Y.~Zhao, D.~Liu, Y.~Mu, M.~Tan, X.~Wang \emph{et~al.}, ``Deep high-resolution representation learning for visual recognition,'' \emph{IEEE Trans. Pattern Anal. Mach. Intell.}, vol.~43, no.~10, pp. 3349--3364, 2020.

\bibitem{zheng2021rethinking}
S.~Zheng, J.~Lu, H.~Zhao, X.~Zhu, Z.~Luo, Y.~Wang, Y.~Fu, J.~Feng, T.~Xiang, P.~H. Torr \emph{et~al.}, ``Rethinking semantic segmentation from a sequence-to-sequence perspective with transformers,'' in \emph{Proc. IEEE Conf. Comput. Vis. Pattern Recognit.}, 2021, pp. 6881--6890.

\bibitem{xie2021segformer}
E.~Xie, W.~Wang, Z.~Yu, A.~Anandkumar, J.~M. Alvarez, and P.~Luo, ``Segformer: Simple and efficient design for semantic segmentation with transformers,'' in \emph{Adv. Neural Inf. Process. Syst.}, vol.~34, 2021, pp. 12\,077--12\,090.

\bibitem{ha2017mfnet}
Q.~Ha, K.~Watanabe, T.~Karasawa, Y.~Ushiku, and T.~Harada, ``Mfnet: Towards real-time semantic segmentation for autonomous vehicles with multi-spectral scenes,'' in \emph{IEEE Int. Conf. Intell. Rob. Syst.}, 2017, pp. 5108--5115.

\bibitem{sun2019rtfnet}
Y.~Sun, W.~Zuo, and M.~Liu, ``Rtfnet: Rgb-thermal fusion network for semantic segmentation of urban scenes,'' \emph{IEEE Robot. Autom. Lett.}, vol.~4, no.~3, pp. 2576--2583, 2019.

\bibitem{shivakumar2020pst900}
S.~S. Shivakumar, N.~Rodrigues, A.~Zhou, I.~D. Miller, V.~Kumar, and C.~J. Taylor, ``Pst900: Rgb-thermal calibration, dataset and segmentation network,'' in \emph{Proc. IEEE Int. Conf. Robot. Autom.}, 2020, pp. 9441--9447.

\bibitem{zhang2021abmdrnet}
Q.~Zhang, S.~Zhao, Y.~Luo, D.~Zhang, N.~Huang, and J.~Han, ``Abmdrnet: Adaptive-weighted bi-directional modality difference reduction network for rgb-t semantic segmentation,'' in \emph{Proc. IEEE Conf. Comput. Vis. Pattern Recognit.}, 2021, pp. 2633--2642.

\bibitem{zhou2021gmnet}
W.~Zhou, J.~Liu, J.~Lei, L.~Yu, and J.-N. Hwang, ``Gmnet: Graded-feature multilabel-learning network for rgb-thermal urban scene semantic segmentation,'' \emph{IEEE Trans. Image Process.}, vol.~30, pp. 7790--7802, 2021.

\bibitem{li2022rgb}
G.~Li, Y.~Wang, Z.~Liu, X.~Zhang, and D.~Zeng, ``Rgb-t semantic segmentation with location, activation, and sharpening,'' \emph{IEEE Trans. Circuit Syst. Video Technol.}, vol.~33, no.~3, pp. 1223--1235, 2022.

\bibitem{zhou2023mmsmcnet}
W.~Zhou, H.~Zhang, W.~Yan, and W.~Lin, ``Mmsmcnet: Modal memory sharing and morphological complementary networks for rgb-t urban scene semantic segmentation,'' \emph{IEEE Trans. Circuit Syst. Video Technol.}, vol.~33, no.~12, pp. 7096--7108, 2023.

\bibitem{dong2023egfnet}
S.~Dong, W.~Zhou, C.~Xu, and W.~Yan, ``Egfnet: Edge-aware guidance fusion network for rgb--thermal urban scene parsing,'' \emph{IEEE Trans. Intell. Transport. Syst.}, 2023.

\bibitem{li2024roadformer}
J.~Li, Y.~Zhang, P.~Yun, G.~Zhou, Q.~Chen, and R.~Fan, ``Roadformer: Duplex transformer for rgb-normal semantic road scene parsing,'' \emph{IEEE Trans. Intell. Veh.}, vol.~9, no.~7, pp. 5163--5172, 2024.

\bibitem{huang2024roadformer+}
J.~Huang, J.~Li, N.~Jia, Y.~Sun, C.~Liu, Q.~Chen, and R.~Fan, ``Roadformer+: Delivering rgb-x scene parsing through scale-aware information decoupling and advanced heterogeneous feature fusion,'' \emph{IEEE Trans. Intell. Veh.}, pp. 1--10, 2024.

\bibitem{zhou2021wsl}
Z.~Helong, S.~Liangchen, C.~Jiajie, Z.~Ye, W.~Guoli, Y.~Junsong, and Q.~Zhang, ``Rethinking soft labels for knowledge distillation: a bias-variance tradeoff perspective,'' in \emph{Int. Conf. Learn. Represent.}, 2021.

\bibitem{bachmann2022multimae}
R.~Bachmann, D.~Mizrahi, A.~Atanov, and A.~Zamir, ``Multimae: Multi-modal multi-task masked autoencoders,'' in \emph{Proc. Eur. Conf. Comput. Vis.}, 2022, pp. 348--367.

\bibitem{he2016deep}
K.~He, X.~Zhang, S.~Ren, and J.~Sun, ``Deep residual learning for image recognition,'' in \emph{Proc. IEEE Conf. Comput. Vis. Pattern Recog.}, 2016, pp. 770--778.

\bibitem{liang2023explicit}
M.~Liang, J.~Hu, C.~Bao, H.~Feng, F.~Deng, and T.~L. Lam, ``Explicit attention-enhanced fusion for rgb-thermal perception tasks,'' \emph{IEEE Robot. Autom. Lett.}, vol.~8, no.~7, pp. 4060--4067, 2023.

\bibitem{zhou2022pan}
M.~Zhou, J.~Huang, Y.~Fang, X.~Fu, and A.~Liu, ``Pan-sharpening with customized transformer and invertible neural network,'' in \emph{Proc. AAAI Conf. Artif. Intell.}, vol.~36, no.~3, 2022, pp. 3553--3561.

\bibitem{cheng2021per}
B.~Cheng, A.~Schwing, and A.~Kirillov, ``Per-pixel classification is not all you need for semantic segmentation,'' in \emph{Adv. Neural Inf. Process. Syst.}, vol.~34, 2021, pp. 17\,864--17\,875.

\bibitem{lin2017feature}
T.-Y. Lin, P.~Doll{\'a}r, R.~Girshick, K.~He, B.~Hariharan, and S.~Belongie, ``Feature pyramid networks for object detection,'' in \emph{Proc. IEEE Conf. Comput. Vis. Pattern Recognit.}, 2017, pp. 2117--2125.

\bibitem{zhang2024semantic}
X.~Zhang, L.~Wang, L.~Zhao, X.~Li, and S.~Ma, ``A semantic-aware and multi-guided network for infrared-visible image fusion,'' \emph{arXiv preprint arXiv:2407.06159}, 2024.

\bibitem{shrivastava2016training}
A.~Shrivastava, A.~Gupta, and R.~Girshick, ``Training region-based object detectors with online hard example mining,'' in \emph{Proc. IEEE Conf. Comput. Vis. Pattern Recog.}, 2016, pp. 761--769.

\bibitem{zhou2021rethinking}
H.~Zhou, L.~Song, J.~Chen, Y.~Zhou, G.~Wang, J.~Yuan, and Q.~Zhang, ``Rethinking soft labels for knowledge distillation: A bias-variance tradeoff perspective,'' \emph{arXiv preprint arXiv:2102.00650}, 2021.

\bibitem{hao2024primkd}
Z.~Hao, Z.~Xiao, Y.~Luo, J.~Guo, J.~Wang, L.~Shen, and H.~Hu, ``Primkd: Primary modality guided multimodal fusion for rgb-d semantic segmentation,'' in \emph{ACM Int. Conf. Multimedia}, 2024, p. 1943–1951.

\bibitem{qiu2024make}
S.~Qiu, J.~Chen, X.~Li, R.~Wan, X.~Xue, and J.~Pu, ``Make a strong teacher with label assistance: A novel knowledge distillation approach for semantic segmentation,'' in \emph{Proc. Eur. Conf. Comput. Vis.}\hskip 1em plus 0.5em minus 0.4em\relax Springer, 2024, pp. 371--388.

\bibitem{ban2024fair}
H.~Ban and K.~Ji, ``Fair resource allocation in multi-task learning,'' \emph{arXiv preprint arXiv:2402.15638}, 2024.

\bibitem{liu2019end}
S.~Liu, E.~Johns, and A.~J. Davison, ``End-to-end multi-task learning with attention,'' in \emph{Proc. IEEE Conf. Comput. Vis. Pattern Recognit.}, 2019, pp. 1871--1880.

\bibitem{guo2022segnext}
M.-H. Guo, C.-Z. Lu, Q.~Hou, Z.~Liu, M.-M. Cheng, and S.-M. Hu, ``Segnext: Rethinking convolutional attention design for semantic segmentation,'' in \emph{Adv. Neural Inf. Process. Syst.}, vol.~35, 2022, pp. 1140--1156.

\bibitem{zheng2024probing}
N.~Zheng, M.~Zhou, J.~Huang, J.~Hou, H.~Li, Y.~Xu, and F.~Zhao, ``Probing synergistic high-order interaction in infrared and visible image fusion,'' in \emph{Proc. IEEE Conf. Comput. Vis. Pattern Recognit.}, 2024, pp. 26\,384--26\,395.

\bibitem{hu2019acnet}
X.~Hu, K.~Yang, L.~Fei, and K.~Wang, ``Acnet: Attention-based network to exploit complementary features for rgb-d semantic segmentation,'' in \emph{Proc. IEEE Int. Conf. Inf. Process.}, 2019, pp. 1440--1444.

\bibitem{chen2020bi}
X.~Chen, K.-Y. Lin, J.~Wang, W.~Wu, C.~Qian, H.~Li, and G.~Zeng, ``Bi-directional cross-modality feature propagation with separation-and-aggregation gate for rgb-d semantic segmentation,'' in \emph{Proc. Eur. Conf. Comput. Vis.}, 2020, pp. 561--577.

\bibitem{zhou2021mffenet}
W.~Zhou, X.~Lin, J.~Lei, L.~Yu, and J.-N. Hwang, ``Mffenet: Multiscale feature fusion and enhancement network for rgb--thermal urban road scene parsing,'' \emph{IEEE Trans. Multimedia}, vol.~24, pp. 2526--2538, 2021.

\bibitem{zhou2022mtanet}
W.~Zhou, S.~Dong, J.~Lei, and L.~Yu, ``Mtanet: Multitask-aware network with hierarchical multimodal fusion for rgb-t urban scene understanding,'' \emph{IEEE Trans. Intell. Veh.}, vol.~8, no.~1, pp. 48--58, 2022.

\bibitem{zhao2022feature}
S.~Zhao and Q.~Zhang, ``A feature divide-and-conquer network for rgb-t semantic segmentation,'' \emph{IEEE Trans. Circuit Syst. Video Technol.}, vol.~33, no.~6, pp. 2892--2905, 2022.

\bibitem{kendall2018multi}
A.~Kendall, Y.~Gal, and R.~Cipolla, ``Multi-task learning using uncertainty to weigh losses for scene geometry and semantics,'' in \emph{Proc. IEEE Conf. Comput. Vis. Pattern Recognit.}, 2018, pp. 7482--7491.

\bibitem{shim2023feedformer}
J.-h. Shim, H.~Yu, K.~Kong, and S.-J. Kang, ``Feedformer: Revisiting transformer decoder for efficient semantic segmentation,'' in \emph{Proc. AAAI Conf. Artif. Intell.}, vol.~37, no.~2, 2023, pp. 2263--2271.



\end{thebibliography}

\end{document}